\documentclass{article}

	\PassOptionsToPackage{numbers,sort,compress}{natbib}

  \usepackage[main, final]{neurips_2026}

\usepackage[utf8]{inputenc} 
\usepackage[T1]{fontenc}    
\usepackage{hyperref}       
\usepackage{url}            
\usepackage{booktabs}       
\usepackage{amsfonts}       
\usepackage{nicefrac}       
\usepackage{microtype}      
\usepackage{xcolor}         
\usepackage{graphicx}
\usepackage{amsmath}
\usepackage{multirow}
\usepackage{colortbl}
\usepackage{pifont}
\usepackage{enumitem}
\usepackage{float}

\title{Brain-OF: An Omnifunctional Foundation Model for fMRI, EEG and MEG}

\author{
	Hanning Guo$^{1,2}$\quad Hanwen Bi$^{3,4}$\quad Farah Abdellatif$^{1,2}$\quad
	Andrei Galbenus$^{1,2}$\\
	\textbf{Jon. N. Shah$^{1,5,6,7}$\quad
	Abigail Morrison$^{8,2}$\quad
	Jürgen Dammers$^{1,9}$}\thanks{Corresponding author}\\[0.5em]
	$^{1}$INM-4, Forschungszentrum Jülich, Germany \\
	$^{2}$Department of Computer Science 3 - Software Engineering, RWTH Aachen University, Germany \\
	$^{3}$INM-7, Forschungszentrum Jülich, Germany\\
	$^{4}$Institute of Systems Neuroscience, Heinrich Heine University, Germany\\
	$^{5}$Department of Neurology, RWTH Aachen University, Germany\\
	$^{6}$JARA-BRAIN-Translational Medicine, Germany \\
	$^{7}$INM–11, JARA, Forschungszentrum Jülich, Germany\\
	$^{8}$IAS-6, Forschungszentrum Jülich, Germany\\
	$^{9}$Department of Psychiatry, Psychotherapy and Psychosomatics, RWTH Aachen University, Germany \\[0.5em]
	\texttt{\{h.guo, j.dammers\}@fz-juelich.de}
}

\begin{document}

\maketitle

\vspace{-1.5em}
\begin{abstract}
  \vspace{-0.8em}
  Brain foundation models have achieved remarkable advances across a wide range of neuroscience tasks. However, most existing models are limited to a single functional modality, restricting their ability to exploit complementary spatiotemporal dynamics and the collective data scale across different neuroimaging techniques.  
  This limitation largely arises from severe \textbf{semantic heterogeneity} and \textbf{resolution discrepancies} among modalities.
  To address these challenges, we propose \textbf{Brain-OF}, an omnifunctional brain foundation model jointly pretrained on fMRI, EEG and MEG, capable of handling both unimodal and multimodal inputs within a unified framework.
  To reconcile heterogeneous spatiotemporal resolutions, we introduce the \emph{Any-Resolution Neural Signal Sampler}, which projects diverse brain signals into a shared semantic space. To further manage semantic shifts, the Brain-OF backbone integrates DINT attention with a Sparse Mixture of Experts, where shared experts capture modality-invariant representations and routed experts specialize in modality-specific semantics.
  Furthermore, to explicitly internalize the characteristics of neural activity through self-supervised learning, we propose \emph{Masked Temporal-Frequency Modeling}, a dual-domain pretraining objective that jointly reconstructs brain signals in both the time and frequency domains.
  Brain-OF is pretrained on a large-scale corpus comprising around 40 datasets and demonstrates superior performance across diverse downstream tasks, highlighting the benefits of joint multimodal integration and dual-domain pretraining. Our code and pretrained model weights are available on our project page \url{https://hanningguo.github.io/BrainOF}.
\end{abstract}

\vspace{-1.5em}
\section{Introduction}
\vskip -0.1in
A central lesson from the recent success of deep learning is that pretraining on large-scale, high-quality and diverse data facilitates the emergence of general-purpose models that consistently outperform task-specific ones~\cite{kaplan2020scaling}. Following this paradigm, brain foundation models have shown promising generalization across a wide range of neural signal processing tasks, including cognitive decoding, affective computing and neurological disease diagnosis~\cite{jiang2024large, wang2024cbramod, dong2025brain, levy2026dance}.
Despite this progress, the scale of available neuroimaging data remains orders of magnitude smaller than the billion-scale corpora in natural language processing and computer vision. Although open-source large-scale neuroimaging collections are growing, data from any single modality remain scarce for training large models due to high acquisition costs and privacy constraints. Meanwhile, most existing brain foundation models are pretrained on a single modality, restricting both data scale and representational diversity, and failing to exploit the complementary characteristics of different brain signals. To address these limitations, we integrate data from the three primary functional neuroimaging modalities (fMRI, EEG and MEG) to develop an \emph{omnifunctional} brain foundation model. 

A pivotal question, therefore, is how to effectively leverage such heterogeneous multi-modal data to pretrain an omnifunctional brain foundation model. The first challenge lies in the variability introduced by diverse scanning devices, sampling rates, acquisition protocols and recording durations across international sites, leading to significant \textbf{structural variability}. A deeper challenge stems from the fundamental differences in the physical imaging mechanisms of each modality.
Specifically, fMRI detects hemodynamic changes, such as oxygenation, blood volume and flow, to measure the Blood-Oxygen-Level-Dependent (BOLD) signal, offering high spatial resolution but limited temporal fidelity~\cite{logothetis2004nature}. In contrast, EEG captures electrical potentials generated by cortical neurons, providing excellent temporal resolution but poor spatial fidelity~\cite{he2018electrophysiological}. MEG records magnetic fields induced by neuronal electrical activity and shares EEG’s high temporal resolution while providing improved spatial localization, though still inferior to that of fMRI~\cite{wheless2004magnetoencephalography}.
These distinct signal origins result in severe differences in spatiotemporal resolution and \textbf{semantic heterogeneity}. However, this heterogeneity also presents a unique opportunity. 
By integrating these modalities, a unified model can exploit the high-frequency temporal dynamics of EEG and MEG, which are lost in the sluggish hemodynamic response of fMRI, while leveraging the superior spatial specificity of fMRI that is inherently lacking in electrophysiological recordings. Such complementary integration enables a holistic characterization of temporal, spectral, and spatial dynamics, ultimately fostering richer and more transferable representations. A third challenge is the inherently low signal-to-noise ratio of brain signals, which can mislead vanilla attention mechanisms to attend to irrelevant fluctuations rather than biologically meaningful patterns.

On the other hand, the success of large-scale pretraining is fundamentally rooted in the generative learning paradigm, which insists that reconstruction compels models to internalize the underlying data distribution~\cite{radford2018improving, han2021pre}. Brain signals are inherently multidimensional, characterized by complex spatial configurations, rich temporal dynamics and structured frequency compositions. 
Although recent brain foundation models have introduced dual-branch architectures to jointly process temporal and spectral features~\cite{wang2024cbramod, zhou2025csbrain}, frequency domain information is typically treated as a static auxiliary input for passive feature extraction, rather than as a target for active reconstruction. 
Therefore, a specialized pretraining objective is essential to explicitly enforce joint reconstruction across both temporal and frequency domains. By coupling temporal and spectral reconstruction, the model is compelled to capture complementary dynamics and intrinsic semantics of neural signals across heterogeneous modalities.

To address these challenges, we propose \textbf{Brain-OF}, an omnifunctional brain foundation model designed to learn holistic representations from neural signals across the three primary functional modalities: fMRI, EEG and MEG.
Our architecture integrates a Sparse Mixture of Experts (MoE) with DINT attention to effectively model heterogeneous semantic tokens generated by the Any-Resolution Neural Signal Sampler (ARNESS), which projects signals of varying spatiotemporal resolutions into a unified semantic space.
The main contributions are summarized as follows:

\begin{itemize}[itemsep=0pt, topsep=0pt, labelsep=4pt, leftmargin=15pt]
	\item \textbf{Unified Omnifunctional Framework.} We introduce Brain-OF, a versatile framework capable of processing both unimodal and multimodal brain data within a single model. To support large-scale pretraining, we curate a corpus of around 40 public datasets spanning fMRI, EEG, and MEG. By cohesively integrating these three complementary neuroimaging modalities, Brain-OF enhances generalization across modalities, tasks, sites and subjects. 
	\item \textbf{Heterogeneity Aligned Architecture.} We propose ARNESS to project heterogeneous neural signals into a shared semantic space. To accommodate semantic diversity, we employ a Sparse MoE architecture in which shared experts capture modality-invariant features while routed experts specialize in modality-specific semantics. Furthermore, we incorporate DINT attention to suppress irrelevant context and efficiently capture both local and global dependencies.
	\item \textbf{Joint Time–Frequency Pretraining.} We introduce Masked Temporal-Frequency Modeling (MTFM), a novel pretraining objective that enforces the joint reconstruction of original signals in both the time and frequency domains. This dual-domain paradigm enables the model to internalize complementary physical dynamics of brain activity, thereby enriching its unified representations. 
\end{itemize}

\vspace{-0.7em}
\section{Related Work}
\vskip -0.1in
\textbf{Brain Foundation Models.}
Foundation models have significantly advanced brain AI by learning generalizable representations from large-scale neuroimaging data. In fMRI, pioneering works such as BrainLM \cite{caro2023brainlm} and BrainMass \cite{yang2024brainmass} employ self-supervised learning to model BOLD time series. Subsequent approaches like Brain-JEPA \cite{dong2024brain} further advanced this direction by employing joint-embedding predictive architectures to improve spatiotemporal representation learning. Parallel advancements in EEG have led to models such as BENDR \cite{kostas2021bendr}, MMM \cite{yi2023learning} and BrainBERT \cite{wang2023brainbert}, which demonstrated the efficacy of contrastive and masked modeling objectives for capturing temporal dynamics in EEG signals. Recently, LaBraM \cite{jiang2024large} and CbraMod \cite{wang2024cbramod} have utilized massive data to enhance general-purpose EEG representation learning. Despite these advances, those foundation models remain confined to single modalities, failing to leverage the complementary characteristics across functional neuroimaging modalities.

\vspace{-0.2em}
\textbf{Heterogeneous Foundation Models.} 
More recently, a growing line of work has begun to address heterogeneity by jointly modeling multiple signal types within unified foundation frameworks. BIOT \cite{yang2023biot} pioneered this direction by pretraining on biosignals, specifically EEG and ECG. Within the electrophysiological domain, Brainwave \cite{yuan2024brainwave} and PopT \cite{chau2025population} learns joint representations across variable channel configurations for both intracranial and scalp EEG, while BrainOmni \cite{xiao2025brainomni} extends to jointly model EEG and MEG signals. Beyond purely electrophysiological modalities, BrainHarmonix \cite{dong2025brain} bridges structural and functional neuroimaging by combining fMRI and MRI. Similarly, PhysioOmni \cite{jiang2025towards} decouples homogeneous and heterogeneous features across multimodal physiological signals, including EEG, ECG, EOG and EMG. 
However, these approaches fall short of capturing the full spectrum of brain activity, missing the opportunity to harmonize the high spatial fidelity of hemodynamic signals with the high temporal resolution of electromagnetic fields. This gap motivates a unified framework for integrating heterogeneous brain signals across fundamentally different physical measurement principles.

\vspace{-0.5em}
\section{Method}
\vskip -0.1in
In this work, we propose \textbf{Brain-OF}, a heterogeneous pretraining framework (Figure~\ref{fig:architecture}) designed as a unified brain foundation model that supports both unimodal and multimodal data. The framework is explicitly designed to achieve broad generalization across four key dimensions: \emph{cross-modality}, \emph{cross-task}, \emph{cross-site} and \emph{cross-subject} settings.
To address the intrinsic heterogeneity of brain signals, Brain-OF introduces an \emph{Any-Resolution Neural Signal Sampler}, which projects inputs with diverse spatial, temporal, and spectral resolutions into a shared semantic space. To further model heterogeneous semantics while mitigating the allocation of attention to irrelevant context, the backbone integrates \emph{DINT attention} with a \emph{Sparse Mixture of Experts}. Finally, to learn informative representations, we propose \emph{Masked Temporal-Frequency Modeling}, which jointly captures complementary knowledge in both the temporal and frequency domains.

\vspace{-0.2em}
\subsection{Model Architecture}
\vskip -0.1in
In this section, the set of modalities is defined as  $\mathcal{M} = \{M_{\text{fMRI}}, M_{\text{EEG}}, M_{\text{MEG}}\}$. For any given modality $M_i \in \mathcal{M}$, the normalized brain signal is represented as a matrix $S \in \mathbb{R}^{N \times T}$, where $N$ denotes the number of channels (for $\text{M/EEG}$) or regions of interest ($\text{ROIs}$; for $\text{fMRI}$), and $T$ represents the temporal length of the signal.

\vspace{-0.2em}
\textbf{Brain Signal Encoder.}
The primary challenge in handling our heterogeneous corpus, spanning diverse modalities, sites, tasks and devices, is the \textbf{structural variability} in channel/ROI counts $N$ and temporal lengths $T$. To standardize the normalized brain signal $S$, we adopt a patching and flattening strategy using a fixed time window $\mathcal{T}$. Specifically, $S$ is segmented into non-overlapping patches of length $\mathcal{T}$ along the temporal dimension for each channel/ROI, then flattened into a consistent sequence $X_c \in \mathbb{R}^{L \times \mathcal{T}}$, where $L = N \cdot \left\lceil \frac{T}{\mathcal{T}} \right\rceil + \text{padding tokens}$. The sequence is padded to a maximum length of 2,048 to ensure uniform input across modalities. Each patch in $X_c$ is subsequently processed by a temporal encoder comprising stacked 1D convolutional layers to capture local temporal dynamics, followed by RMSNorm and GELU activation. This yields a sequence of patch embeddings $X \in \mathbb{R}^{L \times D}$, where $D$ is the embedding dimension:
\[
X = \{x_{\text{patch}_1}, x_{\text{patch}_2}, \dots, x_{\text{patch}_{N \cdot \lceil T/\mathcal{T} \rceil}}, \dots, x_{\text{padding}_L} \}
\]
An attention mask is generated to exclude zero-padded patches from attention computations, eliminating spurious interactions and ensuring training stability.

\begin{figure*}[ht]
	\vskip -0.05in
	\centering
	\includegraphics[width=0.99\linewidth]{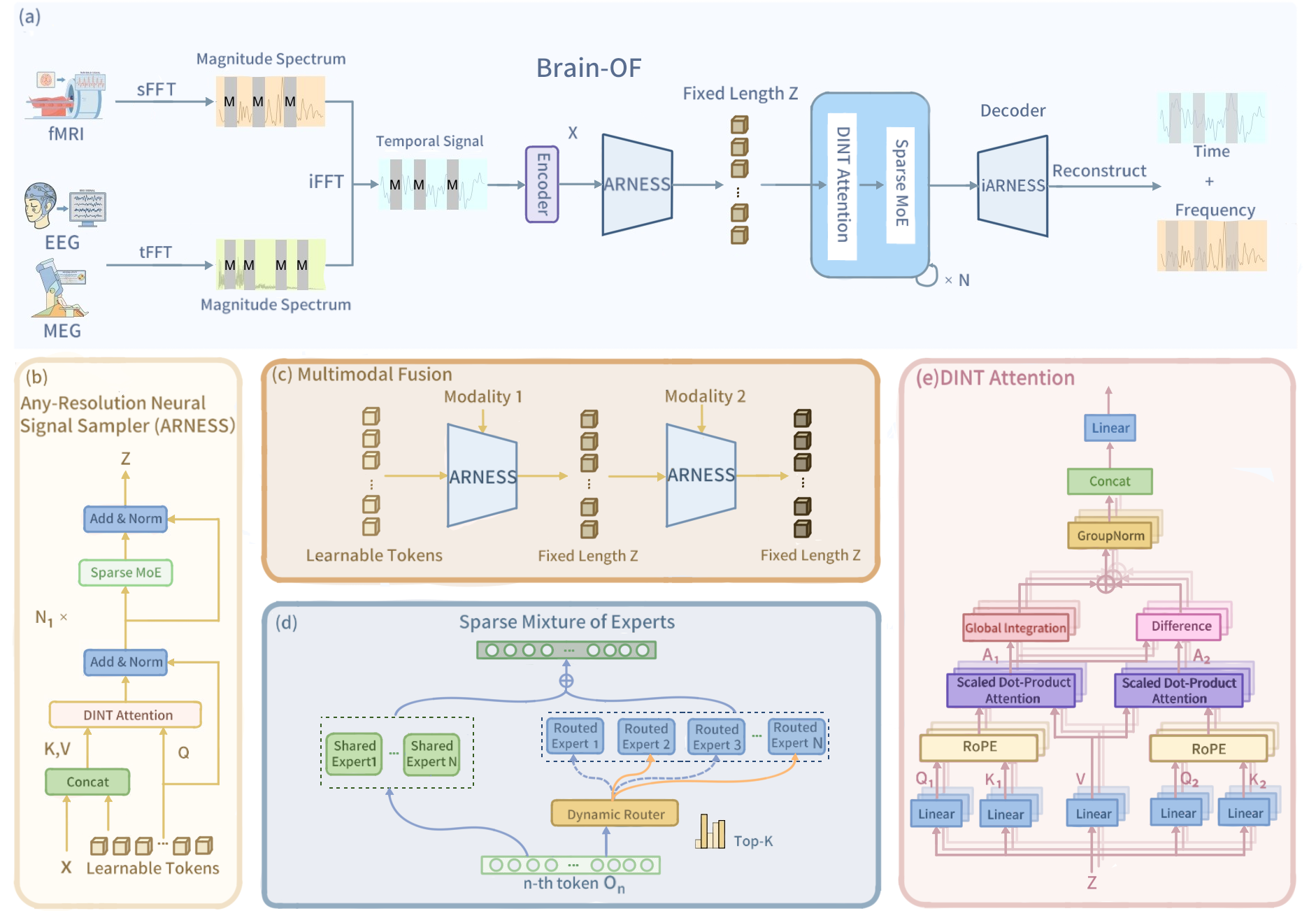}
	\vskip -0.1in
	\caption{
		\textbf{Overview of the Brain-OF Architecture.} 
		Brain-OF is an omnifunctional foundation model jointly pretrained on fMRI, EEG and MEG.
		\textbf{(a) Pretraining pipeline:} The original signals are masked in both the temporal and frequency domains to encourage the model to jointly learn coupled time–frequency representations.
		\textbf{(b) ARNESS} projects arbitrary resolution signals into a unified semantic space and also serves as a decoder to inverse-sample latent representations for reconstruction during pretraining.
		\textbf{(c) Multimodal Fusion} serially resamples multiple unimodal sequences via ARNESS without introducing modality-specific fusion branches.
		\textbf{(d) Sparse MoE} addresses semantic heterogeneity by dynamically routing tokens to specialized experts while extracting modality-invariant knowledge through shared experts.
		\textbf{(e) DINT Attention} mitigates attention allocation to irrelevant contextual information.
		Here, sFFT and tFFT denote Fast Fourier Transform along spatial and temporal dimensions; iFFT denotes the inverse Fast Fourier Transform; and \textit{M} indicates the masking.
	}
	\label{fig:architecture}
	\vskip -0.1in
\end{figure*} 

\textbf{Any-Resolution Neural Signal Sampler.}
Despite the structural unification of the input into sequence $X$, intrinsic differences in temporal, spatial and spectral resolutions persist across modalities, resulting in significant \textbf{semantic heterogeneity}. To bridge this gap and enable a single model backbone to efficiently process diverse unimodal or multimodal signals, we introduce the \textbf{A}ny-\textbf{R}esolution \textbf{Ne}ural \textbf{S}ignal \textbf{S}ampler (ARNESS). As illustrated in Figure~\ref{fig:architecture}b, ARNESS employs Perceiver-style cross-attention~\cite{alayrac2022flamingo} to project the heterogeneous sequence $X$ into a shared semantic space via a fixed number of learnable latent tokens $Z \in \mathcal{R}^{C \times D}$ ($C=128$ in our implementation). Specifically, $X$ and $Z$ are concatenated to form the Keys and Values, with $Z$ serving as the Queries in cross-attention, thereby resampling essential semantic information from the high-dimensional $X$ into a compact, fixed-length $Z$:
\[
ARNESS(X,Z)=CrossAttn(Q=Z,KV=[X;Z],Attn\_Mask)
\] 
This approach enables the subsequent backbone to handle all inputs uniformly, despite inherent heterogeneity.
Lower-resolution modalities can potentially benefit by preserving the meaningful semantic space shaped by higher-resolution modalities.
It also dramatically reduces the number of tokens for high-dimension brain signals, significantly lowering pretraining computational costs.

Moreover, ARNESS naturally supports multimodal fusion (Figure~\ref{fig:architecture}c) by sequentially resampling multiple unimodal sequences into a single fixed-length representation \textbf{without requiring modality-specific fusion branches}:
\[
Z^{(i+1)} = ARNESS(X^{(i)}, Z^{(i)}),\ \ \ i \in \mathcal{M}
\]

\textbf{DINT Attention Mechanism.}
The vanilla self-attention mechanism often overallocates attention to irrelevant context, resulting in "attention noise." This issue is particularly detrimental in neuroimaging, where residual noise in preprocessed brain signals can exacerbate this phenomenon, leading to coarse attention patterns and degraded performance~\cite{guo2025xaiguiformer}. To mitigate this, we introduce the \textbf{D}ifferential-\textbf{Int}egral (DINT) Attention mechanism~\cite{cang2025dint}, which extends Differential Attention~\cite{ye2024differential} to capture precise local dependencies while amplifying global relevance and suppressing noise in the semantic space $Z$.

The differential component reduces noise by computing the difference between two attention distributions. The query and key projections from $Z$ are split into halves, $(Q_1, Q_2)$ and $(K_1, K_2)$, respectively. The differential attention matrix $A_{\text{diff}}$ is defined as:
\[A_{\text{diff}} = \text{softmax}\left(\frac{Q_1 K_1^\top}{\sqrt{d}}\right) - \lambda \cdot \text{softmax}\left(\frac{Q_2 K_2^\top}{\sqrt{d}}\right)\]
where the learnable scalar $\lambda$ is reparameterized as $\lambda = \exp(\lambda_{q_1} \cdot \lambda_{k_1}) - \exp(\lambda_{q_2} \cdot \lambda_{k_2}) + \lambda_{\text{init}}$. $\lambda_{\text{init}}$ is a constant initialization value and $\lambda_{q,k}$ are learnable parameters. Following empirical settings~\cite{ye2024differential}, we set $\lambda_{\text{init}} = 0.8 - 0.6 \cdot \exp(-0.3 \cdot (l - 1))$, where $l$ denotes the layer depth.

To capture global dependencies, the integral component computes a global importance score by averaging the columns of the first attention map $A_1$ (from $Q_1, K_1$):
\[G = \frac{1}{C} \sum_{i=1}^{C} A_1[i, j]\]
yielding $G \in \mathbb{R}^{1 \times C}$ (\(C\) denotes the number of tokens, with \(C=128\) in the backbone). To align dimensions for combination, $G$ is expanded along the query dimension to form $G_{\text{expanded}} \in \mathbb{R}^{C \times C}$.
The final output integrates local (differential) and global (integral) semantics:
\[\text{DINTAttn}(Z) = \left( A_{\text{diff}} + \lambda \cdot G_{\text{expanded}} \right) V\]
In addition, $G_{\text{expanded}}$ ensures row sums of one, guaranteeing numerical stability. Rotary Position Embeddings (RoPE)~\cite{su2024roformer} are used in our model to encode relative positional information within the sequence.

\textbf{Sparse Mixture of Experts.}
Our preliminary experiments using dense feedforward layers revealed a \emph{modality seesaw} effect, where improvements for one modality (e.g., fMRI) degraded performance for others (EEG/MEG). This arises from the diverse semantic features of fMRI, EEG and MEG within the sequence $Z$. To address this, we adopt a \textbf{S}parse \textbf{M}ixture \textbf{o}f \textbf{E}xperts (SMoE) architecture, which excels at modeling heterogeneous semantic spaces. Specifically, we employ a combination of shared experts for modality-invariant representations and routed experts for modality-specific information. Each expert is implemented via a Feed-Forward Network (FFN). For the $n$-th token $O_n$ from the DINT attention output, the SMoE layer is defined as:
\[\text{SMoE}(O_n) = O_n + \sum_{i=1}^{N_s} \text{FFN}_i^{(s)}(O_n) + \sum_{i \in \mathcal{I}_n} g_{i,n} \text{FFN}_i^{(r)}(O_n)\]
where $N_s$ is the number of shared experts, and $\mathcal{I}_n$ denotes the indices of the top-$k$ routed experts selected for token $O_n$. The routing weights $g_{i,n}$ are normalized over the selected routed experts: 
\[g_{i,n} = \frac{p_{i,n}}{\sum_{j \in \mathcal{I}_n} p_{j,n}}\]
where $p_{i,n}$ is the raw probability generated by a sigmoid-activated gating network. To prevent expert collapse, we use an auxiliary-loss-free load-balancing strategy~\cite{wang2024auxiliary}. A bias term $b_i$ is added to $p_{i,n}$ prior to top-$k$ selection (i.e., selection is based on $p_{i,n} + b_i$), while the actual routing weights for aggregating routed experts rely solely on $p_{i,n}$. At the end of each training step, $b_i$ is dynamically adjusted by a bias update rate $\gamma$: decreased for overloaded experts and increased for underloaded ones. Through the dynamic adjustment, SMoE keeps balanced expert load without additional auxiliary losses.

\vspace{-0.5em}
\subsection{Pretraining}
\vskip -0.1in
\textbf{Masked Temporal-Frequency Modeling (MTFM).}
The efficacy of foundation models stems largely from self-supervised pretraining, which facilitates the learning of rich representations by reconstructing masked inputs to capture underlying data distributions. In neuroimaging, brain signals inherently contain complementary information across the temporal, spatial and frequency domains. Consequently, effective representation learning requires a model to internalize coupled features from both spectral and temporal perspectives.
To achieve this, we introduce \textbf{M}asked \textbf{T}emporal–\textbf{F}requency \textbf{M}odeling (MTFM) for learning informative representations across temporal and frequency domains in all modalities. Given a normalized timecourse of a  brain signal $S$, we first transform it into its frequency domain representation $F_S$. Due to the distinct characteristics of different modalities, richer spatial information in fMRI and higher temporal resolution in EEG and MEG, we apply modality-specific frequency transformations:
\[F_S =
\begin{cases}
	\mathcal{F}_{\text{spatial}}(S), & \text{if } S \in M_{\text{fMRI}} \\
	\mathcal{F}_{\text{temporal}}(S), & \text{if } S \in \{ M_{\text{EEG}}, M_{\text{MEG}} \}
\end{cases}\]
where $\mathcal{F}$ denotes the Fourier transform applied along the spatial dimension for fMRI or the temporal dimension for EEG and MEG. Here, the spatial FFT over ROIs is not intended to model physical spatial frequencies, but serves as a fixed orthogonal transform to encourage global cross-region structure learning. Within the frequency domain, we apply a random frequency mask $M_F$, with a masking ratio sampled from a normal distribution. The masked frequency representation is then transformed back to the original signal domain via the inverse Fourier transform. To further enhance temporal modeling, a second masking operation $M_T$ is applied to randomly occlude temporal patches in this spectrally perturbed signal. Following the backbone, a lightweight one-layer ARNESS decoder inverse-samples the latent representations to reconstruct the signal. The model is pretrained to jointly reconstruct both the masked temporal patches and the frequency components, thereby fostering the learning of coupled time–frequency representations.

\textbf{Loss Function.}
The pretraining objective combines reconstruction losses in both the temporal and frequency domains:
\[
\mathcal{L}_{\text{MTFM}} = (1 - \alpha) \cdot \mathcal{L}_{\text{SL1}}(\hat{S}, S) + \alpha \cdot \mathcal{L}_{\text{SL1}}(|\hat{F}_S|, |F_S|)
\]
where $\hat{S}$ denotes the reconstructed signal and $\hat{F}_S = \mathcal{F}(\hat{S})$ is its frequency domain representation; $|\cdot|$ indicates the magnitude; and $\alpha=0.8$ is empirically chosen to balance the temporal and frequency reconstruction terms. We utilize the smooth L1 loss ($\mathcal{L}_{\text{SL1}}$) to ensure stable optimization and robustness to outliers.
\vskip -0.1in

\begin{table*}[t]
	\caption{Overview of downstream tasks and evaluation datasets.}
	\label{downstream_datasets}
	\centering
	\begin{small}
		\setlength{\tabcolsep}{2.2pt}
		\fontsize{9pt}{9pt}\selectfont
		\begin{tabular}{lccccccr}
			\toprule
			\textbf{Downstream Tasks} & \textbf{Datasets} & \textbf{Modalities} & \textbf{\#Chs/ROIs} & \textbf{Durations} & \textbf{\#Subjects} & \textbf{\#Samples} & \textbf{Labels}  \\
			\midrule
			Emotion Recognition & SEED-V & EEG & 62 & 1s & 20 & 117,744 & 5-class  \\
			Seizure Detection & CHB-MIT & EEG & 22 & 10s & 22 & 326,993 & 2-class  \\
			Abnormal Detection & TUAB & EEG & 19 & 10s & 2,383 & 409,455 & 2-class  \\
			AD Diagnosis & ADNI & fMRI & 246 & $\sim$300s & 222 & 618 & 2-class  \\
			ADHD Identification & ADHD-200 & fMRI & 246 & $\sim$360s & 926 & 1,369 & 2-class  \\
			\multirow{3}{*}{Brain Age Prediction}
			& CamCAN & fMRI & 246 & 520s & 652 & 652 & regression  \\
			& CamCAN & MEG & 306 & 2s & 643 & 177,044 & regression  \\
			& CamCAN & MEG + fMRI & - & - & 621 & 171,057 & regression  \\
			\multirow{3}{*}{Aging Classification}
			& LEMON & fMRI & 246 & 448s & 219 & 438 & 2-class  \\
			& LEMON & EEG & 61 & 10s & 203 & 17,235 & 2-class  \\
			& LEMON & EEG + fMRI & - & - & 194 & 16,471 & 2-class  \\
			\bottomrule
		\end{tabular}
	\end{small}
	\vskip -0.2in
\end{table*}


\section{Experiments}

\vskip -0.1in
\subsection{Datasets}
\vskip -0.05in
\textbf{Pre-training.} We curate a heterogeneous collection of \textbf{37 public neuroimaging datasets} to jointly pretrain Brain-OF across fMRI, EEG and MEG modalities. In total, the pretraining dataset comprises \textbf{32,278 unique participants and 5,870,984 samples}. This diverse dataset integrates a wide range of M/EEG channel configurations and both 3T and 7T fMRI recordings from multiracial, multicultural international sites. Such heterogeneity in acquisition protocols, populations and modalities enables the construction of a general-purpose, omnifunctional brain foundation model.
Further details on dataset descriptions and preprocessing are provided in Appendix~\ref{detailsofDatasets} and Table~\ref{pretraining_datasets}.

\begin{table*}[t]
	\caption{Comparison results of different methods on downstream tasks.}
	\label{Comparison_results}
	\centering
	\setlength{\tabcolsep}{1.1pt}
		\fontsize{8pt}{9pt}\selectfont
		\hspace*{-0.05in}
		\begin{tabular}{l@{\hspace{0.5pt}}ccccccccc}
			\toprule
			
			\multirow{5}{*}{\textbf{Methods}} & \textbf{SEED-V} & \textbf{CHB-MIT} & \textbf{TUAB} & \textbf{ADNI} & \textbf{ADHD-200} & \textbf{CamCAN} & \textbf{CamCAN} & \textbf{LEMON} & \textbf{LEMON}  \\
			\cmidrule(r{0.02in}){2-10}
			& \textbf{EEG} & \textbf{EEG} & \textbf{EEG} & \textbf{fMRI} & \textbf{fMRI} & \textbf{fMRI} & \textbf{MEG} & \textbf{fMRI} & \textbf{EEG} \\
			\cmidrule(r{0.02in}){2-10}
			& Kappa(\%) $\uparrow$ & BAC(\%) $\uparrow$ & BAC(\%) $\uparrow$ & BAC(\%) $\uparrow$ & BAC(\%) $\uparrow$ & MAE $\downarrow$ & MAE $\downarrow$ & BAC(\%) $\uparrow$ & BAC(\%) $\uparrow$ \\
			\midrule
			EEGNet &
			10.06{\fontsize{5pt}{0pt}\selectfont$\pm$1.43} &
			56.58{\fontsize{5pt}{0pt}\selectfont$\pm$1.06} &
			76.42{\fontsize{5pt}{0pt}\selectfont$\pm$0.36} &
			60.22{\fontsize{5pt}{0pt}\selectfont$\pm$2.01} &
			57.66{\fontsize{5pt}{0pt}\selectfont$\pm$2.03} &
			15.57{\fontsize{5pt}{0pt}\selectfont$\pm$1.79} &
			11.60{\fontsize{5pt}{0pt}\selectfont$\pm$0.47} &
			59.12{\fontsize{5pt}{0pt}\selectfont$\pm$2.10} &
			78.84{\fontsize{5pt}{0pt}\selectfont$\pm$2.89} 
			\\
			EEGConformer &
			17.72{\fontsize{5pt}{0pt}\selectfont$\pm$1.74} &
			59.76{\fontsize{5pt}{0pt}\selectfont$\pm$1.41} &
			77.58{\fontsize{5pt}{0pt}\selectfont$\pm$0.49} &
			65.16{\fontsize{5pt}{0pt}\selectfont$\pm$2.45} &
			62.00{\fontsize{5pt}{0pt}\selectfont$\pm$2.64} &
			\underline{8.56{\fontsize{5pt}{0pt}\selectfont$\pm$0.72}} &
			10.03{\fontsize{5pt}{0pt}\selectfont$\pm$0.21} & 67.27{\fontsize{5pt}{0pt}\selectfont$\pm$4.15}
			& 78.94{\fontsize{5pt}{0pt}\selectfont$\pm$2.35}
			\\
			ContraWR & 
			19.05{\fontsize{5pt}{0pt}\selectfont$\pm$1.88} & 63.44{\fontsize{5pt}{0pt}\selectfont$\pm$0.02} & 77.46{\fontsize{5pt}{0pt}\selectfont$\pm$0.41} & 59.69{\fontsize{5pt}{0pt}\selectfont$\pm$4.97} & 62.67{\fontsize{5pt}{0pt}\selectfont$\pm$1.69} & 9.15{\fontsize{5pt}{0pt}\selectfont$\pm$0.39}  & 9.01{\fontsize{5pt}{0pt}\selectfont$\pm$0.40} &
			63.97{\fontsize{5pt}{0pt}\selectfont$\pm$1.28} & 77.42{\fontsize{5pt}{0pt}\selectfont$\pm$0.45}   \\
			CNN-Transformer & 
			20.72{\fontsize{5pt}{0pt}\selectfont$\pm$1.83} & 63.89{\fontsize{5pt}{0pt}\selectfont$\pm$0.67} & 77.77{\fontsize{5pt}{0pt}\selectfont$\pm$0.22} & 62.27{\fontsize{5pt}{0pt}\selectfont$\pm$4.87} & 61.94{\fontsize{5pt}{0pt}\selectfont$\pm$0.43} & 16.00{\fontsize{5pt}{0pt}\selectfont$\pm$0.10} & 15.69{\fontsize{5pt}{0pt}\selectfont$\pm$0.09}& 64.83{\fontsize{5pt}{0pt}\selectfont$\pm$1.54} & 78.68{\fontsize{5pt}{0pt}\selectfont$\pm$0.77}   \\
			FFCL & 
			20.78{\fontsize{5pt}{0pt}\selectfont$\pm$2.01} & 62.62{\fontsize{5pt}{0pt}\selectfont$\pm$1.04} & 78.48{\fontsize{5pt}{0pt}\selectfont$\pm$0.38} & 60.31{\fontsize{5pt}{0pt}\selectfont$\pm$2.92} & 63.20{\fontsize{5pt}{0pt}\selectfont$\pm$3.96} & \textbf{8.50{\fontsize{5pt}{0pt}\selectfont$\pm$1.24}} & 8.83{\fontsize{5pt}{0pt}\selectfont$\pm$0.28}& 62.93{\fontsize{5pt}{0pt}\selectfont$\pm$2.86} & \underline{80.17{\fontsize{5pt}{0pt}\selectfont$\pm$1.89}}   \\
			ST-Transformer & 
			10.83{\fontsize{5pt}{0pt}\selectfont$\pm$1.21} & 59.15{\fontsize{5pt}{0pt}\selectfont$\pm$1.95} & 79.66{\fontsize{5pt}{0pt}\selectfont$\pm$0.23} & 59.69{\fontsize{5pt}{0pt}\selectfont$\pm$3.58} & 58.40{\fontsize{5pt}{0pt}\selectfont$\pm$2.33} & 12.39{\fontsize{5pt}{0pt}\selectfont$\pm$0.46} & 10.69{\fontsize{5pt}{0pt}\selectfont$\pm$0.38} & 63.62{\fontsize{5pt}{0pt}\selectfont$\pm$3.01} & 76.95{\fontsize{5pt}{0pt}\selectfont$\pm$1.14} \\
			BrainNetTF &
			12.69{\fontsize{5pt}{0pt}\selectfont$\pm$0.58} &
			50.98{\fontsize{5pt}{0pt}\selectfont$\pm$0.73} &
			75.26{\fontsize{5pt}{0pt}\selectfont$\pm$0.40} &
			\underline{69.62{\fontsize{5pt}{0pt}\selectfont$\pm$2.23}} &
			61.72{\fontsize{5pt}{0pt}\selectfont$\pm$0.76} &
			9.52{\fontsize{5pt}{0pt}\selectfont$\pm$0.37} &
			9.90{\fontsize{5pt}{0pt}\selectfont$\pm$0.06} &
			\underline{69.77{\fontsize{5pt}{0pt}\selectfont$\pm$3.23}} & 60.93{\fontsize{5pt}{0pt}\selectfont$\pm$2.72}
			\\
			BrainNetCNN &
			11.06{\fontsize{5pt}{0pt}\selectfont$\pm$1.04} &
			51.87{\fontsize{5pt}{0pt}\selectfont$\pm$1.85} &
			74.26{\fontsize{5pt}{0pt}\selectfont$\pm$0.87} &
			67.20{\fontsize{5pt}{0pt}\selectfont$\pm$1.58} &
			60.55{\fontsize{5pt}{0pt}\selectfont$\pm$1.95} &
			12.29{\fontsize{5pt}{0pt}\selectfont$\pm$1.29} &
			14.67{\fontsize{5pt}{0pt}\selectfont$\pm$0.14} &
			\textbf{70.08{\fontsize{5pt}{0pt}\selectfont$\pm$1.17}} & 62.01{\fontsize{5pt}{0pt}\selectfont$\pm$0.81}
			\\
			\midrule
			LaBraM & 
			23.86{\fontsize{5pt}{0pt}\selectfont$\pm$2.09} & 70.75{\fontsize{5pt}{0pt}\selectfont$\pm$3.58} & 81.40{\fontsize{5pt}{0pt}\selectfont$\pm$0.19} & 60.26{\fontsize{5pt}{0pt}\selectfont$\pm$2.84} & 56.58{\fontsize{5pt}{0pt}\selectfont$\pm$3.22} & 16.10{\fontsize{5pt}{0pt}\selectfont$\pm$0.04} & 13.66{\fontsize{5pt}{0pt}\selectfont$\pm$1.46} & 56.43{\fontsize{5pt}{0pt}\selectfont$\pm$2.40} & 75.59{\fontsize{5pt}{0pt}\selectfont$\pm$2.60}  \\
			CBraMod & 
			\underline{25.69{\fontsize{5pt}{0pt}\selectfont$\pm$1.43}} & 73.98{\fontsize{5pt}{0pt}\selectfont$\pm$2.84} & 78.91{\fontsize{5pt}{0pt}\selectfont$\pm$0.30} & 61.01{\fontsize{5pt}{0pt}\selectfont$\pm$4.57} & 60.09{\fontsize{5pt}{0pt}\selectfont$\pm$5.66} & 15.80{\fontsize{5pt}{0pt}\selectfont$\pm$0.14} & 15.71{\fontsize{5pt}{0pt}\selectfont$\pm$0.21} & 53.84{\fontsize{5pt}{0pt}\selectfont$\pm$4.40} & 75.29{\fontsize{5pt}{0pt}\selectfont$\pm$6.29}  \\
			BrainHarmonix-F & 
			7.71{\fontsize{5pt}{0pt}\selectfont$\pm$0.92}&  56.08{\fontsize{5pt}{0pt}\selectfont$\pm$0.76}& 77.94{\fontsize{5pt}{0pt}\selectfont$\pm$0.14} & 62.16{\fontsize{5pt}{0pt}\selectfont$\pm$1.15} & 62.78{\fontsize{5pt}{0pt}\selectfont$\pm$2.56} & 14.81{\fontsize{5pt}{0pt}\selectfont$\pm$0.49} & \underline{8.73{\fontsize{5pt}{0pt}\selectfont$\pm$0.09}} & 63.11{\fontsize{5pt}{0pt}\selectfont$\pm$1.32} &77.95{\fontsize{5pt}{0pt}\selectfont$\pm$2.36} \\
			Brain-JEPA &
			3.19{\fontsize{5pt}{0pt}\selectfont$\pm$0.21} &
			71.66{\fontsize{5pt}{0pt}\selectfont$\pm$1.22}&
			74.25{\fontsize{5pt}{0pt}\selectfont$\pm$0.59}&
			61.51{\fontsize{5pt}{0pt}\selectfont$\pm$2.10}&
			60.58{\fontsize{5pt}{0pt}\selectfont$\pm$2.43}&
			12.28{\fontsize{5pt}{0pt}\selectfont$\pm$0.70}&
			11.40{\fontsize{5pt}{0pt}\selectfont$\pm$0.28}&
			63.79{\fontsize{5pt}{0pt}\selectfont$\pm$2.86}&
			72.87{\fontsize{5pt}{0pt}\selectfont$\pm$1.11}
			\\
			\midrule
			Brain-OF Base & 
			23.98{\fontsize{5pt}{0pt}\selectfont $\pm$0.29} & 73.99{\fontsize{5pt}{0pt}\selectfont $\pm$1.48} & 81.88{\fontsize{5pt}{0pt}\selectfont $\pm$0.17} & 68.23{\fontsize{5pt}{0pt}\selectfont$\pm$2.07} & 62.57{\fontsize{5pt}{0pt}\selectfont$\pm$1.19} & 12.09{\fontsize{5pt}{0pt}\selectfont$\pm$0.78} & 8.99{\fontsize{5pt}{0pt}\selectfont$\pm$0.14} & 58.95{\fontsize{5pt}{0pt}\selectfont$\pm$2.13} & 76.05{\fontsize{5pt}{0pt}\selectfont$\pm$0.44} \\
			Brain-OF Large &
			24.68{\fontsize{5pt}{0pt}\selectfont $\pm$0.37} &
			\underline{74.12{\fontsize{5pt}{0pt}\selectfont $\pm$1.13}}&
			\underline{81.90{\fontsize{5pt}{0pt}\selectfont $\pm$0.15}} &
			69.01{\fontsize{5pt}{0pt}\selectfont$\pm$2.62} &
			\underline{63.30{\fontsize{5pt}{0pt}\selectfont$\pm$0.47}} &
			11.71{\fontsize{5pt}{0pt}\selectfont$\pm$0.67} &
			8.86{\fontsize{5pt}{0pt}\selectfont$\pm$0.21} &
			62.85{\fontsize{5pt}{0pt}\selectfont$\pm$4.29} & 78.04{\fontsize{5pt}{0pt}\selectfont$\pm$1.59}
			\\
			Brain-OF Huge &
			\textbf{26.36$^{*}${\fontsize{5pt}{0pt}\selectfont$\pm$0.17}} & \textbf{74.87{\fontsize{5pt}{0pt}\selectfont$\pm$0.84}} & \textbf{82.87$^{*}${\fontsize{5pt}{0pt}\selectfont$\pm$0.18}}&\textbf{71.79{\fontsize{5pt}{0pt}\selectfont$\pm$2.19}} & \textbf{63.51{\fontsize{5pt}{0pt}\selectfont$\pm$1.01}}
			& 9.51{\fontsize{5pt}{0pt}\selectfont$\pm$0.51}& \textbf{7.87$^{*}${\fontsize{5pt}{0pt}\selectfont$\pm$0.22}} & 61.79{\fontsize{5pt}{0pt}\selectfont$\pm$1.29} & \textbf{80.39{\fontsize{5pt}{0pt}\selectfont$\pm$0.59}}
			\\
			\bottomrule
		\end{tabular}
	\begin{minipage}{\linewidth}
		\footnotesize
		Performance is reported as mean $\pm$ standard deviation across five random seeds. The best result is highlighted in bold and the second-best result is $\underline{\text{underlined}}$. $^{*}$ indicates a statistically significant improvement over the best baseline ($p<0.05$).
	\end{minipage}
	\vskip -0.2in
\end{table*}

\vskip -0.01in
\textbf{Downstream Tasks.}
We evaluate Brain-OF on seven downstream tasks across 11 experiments spanning \textbf{fMRI, EEG and MEG} modalities in both unimodal and multimodal settings.
To assess the generalizability of the model on EEG, we include emotion recognition (SEED-V), seizure detection (CHB-MIT) and abnormal detection (TUAB) as downstream tasks.
For fMRI, two large multi-site datasets are used to evaluate diagnostic performance on neurodegenerative Alzheimer’s disease (ADNI) and neurodevelopmental attention deficit hyperactivity disorder (ADHD-200), respectively.
In addition, brain age prediction (CamCAN) serves as a regression benchmark to evaluate unimodal MEG, unimodal fMRI, and multimodal MEG+fMRI fusion. A second multimodal dataset (LEMON) is further employed to assess age-group classification in unimodal EEG, unimodal fMRI, and multimodal EEG+fMRI settings.
Together, these downstream tasks evaluate Brain-OF’s effectiveness on resting-state, task-state and ictal/interictal brain activity, while probing single-modality and multi-modality fusion.
We adopt cross-subject/trial splits following \cite{wang2024cbramod} for EEG datasets and a stratified split following \cite{kan2022brain} for fMRI datasets (7:1.5:1.5), ensuring strict subject-wise separation of training, validation and test sets.
Detailed dataset configurations are provided in Table~\ref{downstream_datasets} and Appendix~\ref{downstreamTaskDatasets}.

\vskip -0.1in
\subsection{Experiment Setups}
\vskip -0.05in
\textbf{Pre-training Settings.} Brain-OF is released in three model variants: a 12-layer Brain-OF Base (47.5M parameters, 21.5M active parameters), a 24-layer Brain-OF Large (331M parameters, 150M active parameters) and a 36-layer Brain-OF Huge (1.7B parameters, 500M active parameters). All models are implemented using Python 3.11.11 and PyTorch 2.4.1+CUDA 12.1.
Model training adopts the AdamW optimizer in BF16 precision with a global batch size of 7,168. The optimization follows a cosine-annealing schedule with a peak learning rate of $3 \times 10^{-3}$, weight decay of $5 \times 10^{-2}$, and optimizer states $\beta=(0.9, 0.95)$ and $\epsilon=10^{-8}$. Variable-length input signals are transformed and padded to a unified sequence length of 2,048. To stabilize routing, we employ an auxiliary-loss-free update for the router bias with coefficient $\gamma = 10^{-3}$.
Pretraining is conducted on up to 56 compute nodes, each equipped with 4 NVIDIA A100 (40GB) GPUs (224 GPUs in total), for approximately six days. Comprehensive hyperparameter configurations are detailed in Appendix~\ref{hyperparameters}.

\vskip -0.05in
\textbf{Baselines.} To provide a comprehensive comparison across modalities, we consider the following 12 baseline models: EEG task-specific models (EEGNet \cite{lawhern2018eegnet}, EEGConformer \cite{song2022eeg}, ContraWR \cite{yang2023self}, CNN-Transformer \cite{peh2022transformer}, FFCL \cite{li2022motor} and ST-Transformer \cite{song2021transformer}), fMRI task-specific models (BrainNetTF \cite{kan2022brain} and BrainNetCNN \cite{kawahara2017brainnetcnn}), EEG foundation models (LaBraM \cite{jiang2024large} and CBraMod \cite{wang2024cbramod}), and fMRI foundation models (BrainHarmonix-F \cite{dong2025brain} and Brain-JEPA \cite{dong2024brain}). Additional details on all baselines are provided in Appendix~\ref{baselines}.

\vskip -0.05in
\textbf{Evaluation Metrics.}
We employ task-specific evaluation metrics for different downstream tasks. For binary classification tasks, we report Balanced Accuracy (BAC), which is also used as the primary metric for model checkpoint selection. In multi-class classification settings, Cohen’s Kappa is adopted as the primary evaluation metric. For regression tasks, performance is assessed using Mean Absolute Error (MAE). All reported results are presented as the mean and standard deviation over five random seeds. For each run, the best model is selected based on the primary metric on the validation set, and final performance is reported on a held-out test set.


\begin{table*}[t]
	\vspace{-1.5em}
	\caption{Ablation studies on downstream tasks.}
	\label{ablation_study}
	\centering
		\setlength{\tabcolsep}{1.4pt}
		
		\fontsize{8pt}{9pt}\selectfont
		\begin{tabular}{l@{\hspace{-3.0pt}}ccccccccc}
			\toprule
			\multirow{5}{*}{\textbf{Methods}} & \textbf{SEED-V} & \textbf{CHB-MIT} & \textbf{TUAB} & \textbf{ADNI} & \textbf{ADHD-200} & \textbf{CamCAN} & \textbf{CamCAN} & \textbf{LEMON} & \textbf{LEMON}  \\
			\cmidrule(r{0.02in}){2-10}
			& \textbf{EEG} & \textbf{EEG} & \textbf{EEG} & \textbf{fMRI} & \textbf{fMRI} & \textbf{fMRI} & \textbf{MEG} & \textbf{fMRI} & \textbf{EEG} \\
			\cmidrule(r{0.02in}){2-10}
			& Kappa(\%) $\uparrow$ & BAC(\%) $\uparrow$ & BAC(\%) $\uparrow$ & BAC(\%) $\uparrow$ & BAC(\%) $\uparrow$ & MAE $\downarrow$ & MAE $\downarrow$ & BAC(\%) $\uparrow$ & BAC(\%) $\uparrow$ \\
			
			\midrule
			Brain-OF (Base) &
			\textbf{23.98}{\fontsize{5pt}{0pt}\selectfont $\pm$0.29} & \textbf{73.99}{\fontsize{5pt}{0pt}\selectfont $\pm$1.48} &
			\textbf{81.88{\fontsize{5pt}{0pt}\selectfont $\pm$0.17}} & \underline{68.23{\fontsize{5pt}{0pt}\selectfont$\pm$2.07}} & \textbf{62.57{\fontsize{5pt}{0pt}\selectfont$\pm$1.19}} & \textbf{12.09{\fontsize{5pt}{0pt}\selectfont$\pm$0.78}} & \underline{8.99{\fontsize{5pt}{0pt}\selectfont$\pm$0.14}} & \underline{58.95{\fontsize{5pt}{0pt}\selectfont$\pm$2.13}} & 76.05{\fontsize{5pt}{0pt}\selectfont$\pm$0.44} \\
			\midrule
			\textit{Objective Ablation} & & & & & & & & & \\
			- w/o MTFM & 
			21.52{\fontsize{5pt}{0pt}\selectfont$\pm$0.57} & 71.43{\fontsize{5pt}{0pt}\selectfont$\pm$2.30} & 81.49{\fontsize{5pt}{0pt}\selectfont$\pm$0.18} & 65.26{\fontsize{5pt}{0pt}\selectfont$\pm$1.83} & \underline{60.35{\fontsize{5pt}{0pt}\selectfont$\pm$1.50}} & 12.80{\fontsize{5pt}{0pt}\selectfont$\pm$1.15} & \textbf{8.96{\fontsize{5pt}{0pt}\selectfont$\pm$0.33}} & 57.18{\fontsize{5pt}{0pt}\selectfont$\pm$1.91} & \textbf{76.90{\fontsize{5pt}{0pt}\selectfont$\pm$0.51}}  \\
			\midrule
			\textit{Modality Ablation} & & & & & & & & & \\
			- w/o fMRI & 
			\underline{22.24{\fontsize{5pt}{0pt}\selectfont$\pm$0.27}} & \underline{71.61{\fontsize{5pt}{0pt}\selectfont$\pm$0.90}} & 80.98{\fontsize{5pt}{0pt}\selectfont$\pm$0.19} & 64.35{\fontsize{5pt}{0pt}\selectfont$\pm$1.77} & 59.94{\fontsize{5pt}{0pt}\selectfont$\pm$1.78} & 12.71{\fontsize{5pt}{0pt}\selectfont$\pm$0.51} & 9.16{\fontsize{5pt}{0pt}\selectfont$\pm$0.13} & 56.89{\fontsize{5pt}{0pt}\selectfont$\pm$3.68} & \underline{76.17{\fontsize{5pt}{0pt}\selectfont$\pm$0.54}}  \\
			- w/o EEG & 
			19.35{\fontsize{5pt}{0pt}\selectfont$\pm$0.40} & 69.07{\fontsize{5pt}{0pt}\selectfont$\pm$1.42} & \underline{81.86{\fontsize{5pt}{0pt}\selectfont$\pm$0.13}} & 67.37{\fontsize{5pt}{0pt}\selectfont$\pm$1.67} & 59.08{\fontsize{5pt}{0pt}\selectfont$\pm$1.15} & 12.30{\fontsize{5pt}{0pt}\selectfont$\pm$0.65} & 9.17{\fontsize{5pt}{0pt}\selectfont$\pm$0.27} & \textbf{60.29{\fontsize{5pt}{0pt}\selectfont$\pm$2.82}} & 75.61{\fontsize{5pt}{0pt}\selectfont$\pm$1.05}  \\
			- w/o MEG & 
			21.62{\fontsize{5pt}{0pt}\selectfont$\pm$0.35} & 70.93{\fontsize{5pt}{0pt}\selectfont$\pm$1.85} & 
			81.50{\fontsize{5pt}{0pt}\selectfont$\pm$0.19} &
			66.68{\fontsize{5pt}{0pt}\selectfont$\pm$2.43} & 60.13{\fontsize{5pt}{0pt}\selectfont$\pm$0.83} & \underline{12.25{\fontsize{5pt}{0pt}\selectfont$\pm$0.97}} & 9.36{\fontsize{5pt}{0pt}\selectfont$\pm$0.15} & 56.68{\fontsize{5pt}{0pt}\selectfont$\pm$1.44} & 75.30{\fontsize{5pt}{0pt}\selectfont$\pm$1.03}  \\
			
			- fMRI-only & 14.69{\fontsize{5pt}{0pt}\selectfont$\pm$0.33} & 67.80{\fontsize{5pt}{0pt}\selectfont$\pm$1.04} & 80.24{\fontsize{5pt}{0pt}\selectfont$\pm$0.19} & \textbf{70.10{\fontsize{5pt}{0pt}\selectfont$\pm$1.64}} & 59.81{\fontsize{5pt}{0pt}\selectfont$\pm$1.11} & 12.36{\fontsize{5pt}{0pt}\selectfont$\pm$0.53} & 9.41{\fontsize{5pt}{0pt}\selectfont$\pm$0.13} & 57.39{\fontsize{5pt}{0pt}\selectfont$\pm$3.06} & 74.48{\fontsize{5pt}{0pt}\selectfont$\pm$0.99} \\
			
			- EEG-only & 
			22.15{\fontsize{5pt}{0pt}\selectfont$\pm$0.53} & 70.15{\fontsize{5pt}{0pt}\selectfont$\pm$1.18} & 79.86{\fontsize{5pt}{0pt}\selectfont$\pm$0.29} & 63.70{\fontsize{5pt}{0pt}\selectfont$\pm$1.79} & 57.01{\fontsize{5pt}{0pt}\selectfont$\pm$1.35} & 12.88{\fontsize{5pt}{0pt}\selectfont$\pm$0.23} & 9.34{\fontsize{5pt}{0pt}\selectfont$\pm$0.09} & 56.37{\fontsize{5pt}{0pt}\selectfont$\pm$1.70} & 75.79{\fontsize{5pt}{0pt}\selectfont$\pm$1.24}\\
			
			- MEG-only &
			20.12{\fontsize{5pt}{0pt}\selectfont$\pm$0.38} & 67.95{\fontsize{5pt}{0pt}\selectfont$\pm$0.93} &
			81.15{\fontsize{5pt}{0pt}\selectfont$\pm$0.11} & 63.46{\fontsize{5pt}{0pt}\selectfont$\pm$2.31} & 57.37{\fontsize{5pt}{0pt}\selectfont$\pm$2.48} & 12.76{\fontsize{5pt}{0pt}\selectfont$\pm$0.73} & 9.05{\fontsize{5pt}{0pt}\selectfont$\pm$0.19} & 57.09{\fontsize{5pt}{0pt}\selectfont$\pm$3.12} & 75.11{\fontsize{5pt}{0pt}\selectfont$\pm$0.53} \\
			
			- No Pretraining & 
			12.63{\fontsize{5pt}{0pt}\selectfont$\pm$1.43} & 63.62{\fontsize{5pt}{0pt}\selectfont$\pm$1.58} & 77.59{\fontsize{5pt}{0pt}\selectfont$\pm$0.27} & 63.39{\fontsize{5pt}{0pt}\selectfont$\pm$2.36} & 56.11{\fontsize{5pt}{0pt}\selectfont$\pm$0.96} & 13.44{\fontsize{5pt}{0pt}\selectfont$\pm$0.85} & 9.69{\fontsize{5pt}{0pt}\selectfont$\pm$1.07} & 54.25{\fontsize{5pt}{0pt}\selectfont$\pm$2.64} & 72.81{\fontsize{5pt}{0pt}\selectfont$\pm$1.38}\\
			\bottomrule
		\end{tabular}
	\vskip -0.2in
\end{table*}

\vspace{-0.5em}
\subsection{Comparison with State-of-the-Art Baselines}
\vskip -0.1in
Brain-OF is evaluated on a diverse set of dowstream tasks including regression, binary and multi-class classification across EEG, fMRI, and MEG modalities.
As detailed in Table~\ref{Comparison_results}, Brain-OF Huge achieves state-of-the-art performance, ranking first in 7 out of 9 tasks with an average rank of 2.2 across all evaluated settings. These results demonstrate its strong cross-modal generalization capability and robustness across heterogeneous neuroimaging modalities.
We observe a consistent performance improvement as model capacity increases. Scaling from Brain-OF Base to Brain-OF Huge yields monotonic improvements across nearly all tasks. Notably, on the CamCAN (MEG) brain-age regression task, the MAE decreases from 8.99 (Base) to 7.87 (Huge), indicating that larger capacity effectively captures fine-grained, continuous biological variations.
An important observation is that task-specific models (e.g., the EEG-based EEGConformer and fMRI-based BrainNetTF) occasionally outperform single-modality foundation models (e.g., the EEG-pretrained LaBraM and fMRI-pretrained BrainHarmonix-F) when tested on modalities different from their pretraining domain. This suggests that foundation models pretrained exclusively on a single modality may over-specialize in modality-specific signal characteristics, leading to limited transferability when applied to other modalities (e.g., EEG-pretrained models on fMRI or MEG tasks). 
In contrast, Brain-OF’s joint multimodal pretraining mitigates this issue by learning shared neural representations while still accommodating modality-specific differences.

\vspace{-0.5em}
\subsection{Ablation Studies}
\vskip -0.1in
To rigorously assess the contributions of the proposed MTFM pretraining objective and the role of heterogeneous modality integration, we conduct comprehensive ablation studies across multiple downstream tasks, as summarized in Table~\ref{ablation_study}.

\vskip -0.01in
\textbf{Effectiveness of MTFM.}
To isolate the benefit of dual-domain pretraining, we compare Brain-OF (pretrained with MTFM) against a baseline pretrained with a standard Masked Autoencoder objective. As shown in Table~\ref{ablation_study}, replacing MTFM with conventional temporal masking consistently degrades performance across most tasks. This confirms that jointly reconstructing signals in both the time and frequency domains forces the model to capture richer neural dynamics than temporal reconstruction alone.
The performance gains are particularly pronounced for tasks requiring complex neural signal decoding, such as SEED-V and CHB-MIT. In contrast, the improvements are relatively modest for more static demographic prediction tasks (e.g., CamCAN and LEMON), where fine-grained frequency dynamics may play a less dominant role.

\vskip -0.01in
\textbf{Cross-Modal Synergy in Joint Pretraining.}
\label{modality_ablation}
We examine the contribution of each modality by selectively removing one or two modalities from the fMRI, EEG and MEG pretraining corpus. As shown in Table~\ref{ablation_study}, the omnifunctional Brain-OF achieves stronger performance than its single- and dual-modality counterparts on most downstream tasks. An exception is observed on ADNI, where the fMRI-only variant slightly outperforms the omnifunctional model. 
This result reflects a potential generalist--specialist trade-off: a modality-specialized model can allocate its full capacity to modality-specific characteristics, whereas Brain-OF is optimized to learn representations that generalize across heterogeneous modalities and tasks. Despite this trade-off, Brain-OF remains competitive on ADNI while achieving more consistent performance across the broader benchmark. Notably, removing the native modality of a downstream task does not always cause a catastrophic performance drop. This suggests that Brain-OF learns robust modality-invariant representations, allowing knowledge transfer across modalities, while modality-specific data still provides additional benefits. Together, these results support the effectiveness of joint cross-modal pretraining beyond unimodal learning.

\begin{figure}[t]
	\centering
	\includegraphics[width=0.99\linewidth]{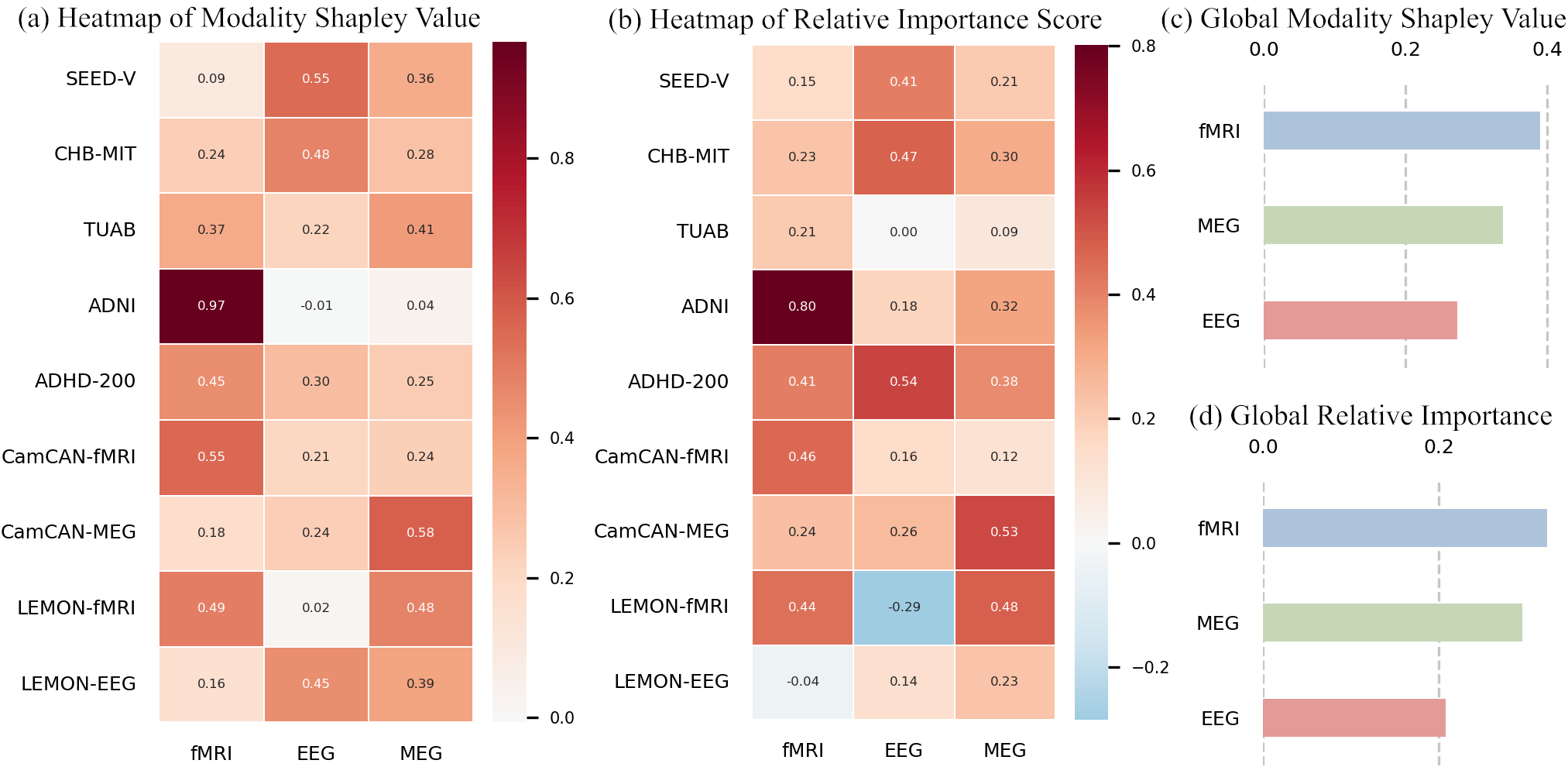}
	\vskip -0.1in
	\caption{\textbf{Modality Importance Analysis.}
	}
	\label{modality_importance}
	
	\vskip -0.3in
\end{figure}

To further quantify how individual modalities contribute under joint pretraining, we employ two complementary Modality Importance Scores introduced in Appendix~\ref{sec:modality}, including the Modality Shapley Value and the Relative Importance Score. We compute both scores for each downstream dataset and further average them across tasks to obtain a global summary, as shown in Figure~\ref{modality_importance}.
The local heatmaps in Figure~\ref{modality_importance}(a) and (b) reveal substantial cross-modal dependencies. While several tasks show strong contributions from their native modality, cross-modal contributions in some cases can be equally important or even larger. For example, on the TUAB EEG benchmark, the Shapley values of MEG (0.41) and fMRI (0.37) both exceed that of EEG (0.22), indicating that pretraining on non-EEG modalities provides transferable information for EEG abnormality detection. Because the Modality Shapley Value measures each modality's marginal contribution across all possible modality coalitions under cooperative game theory, it provides a principled view of such cross-modal synergy. More broadly, most off-modality entries in Figure~\ref{modality_importance}(a) are positive, indicating that modalities generally contribute constructively to downstream tasks outside their own domain, although a few negative values suggest occasional cross-modal interference.
Figure~\ref{modality_importance}(c) and (d) further aggregate these task-level scores into global rankings. Under both measures, fMRI emerges as the most influential modality overall (global Modality Shapley Value: 0.39; global Relative Importance Score: 0.32), followed by MEG (0.34 and 0.29) and EEG (0.27 and 0.21). These results indicate that joint pretraining benefits from both modality-specific information and transferable cross-modal structure, with fMRI providing the strongest overall contribution across the evaluated tasks.

\begin{table*}[b]
	\vskip -0.35in
	\caption{
		\textbf{Performance benefits of multimodal fusion.}
	}
	\label{multimodal}
	\renewcommand{\arraystretch}{0.90}
	\centering
	\begin{small}
		\setlength{\tabcolsep}{9pt} 
		\begin{tabular}{lllccc}
			\toprule
			\multirow{3}{*}{\textbf{Datasets}}& \multirow{3}{*}{\textbf{Metrics}} & \multirow{3}{*}{\textbf{Modality}} & \multicolumn{3}{c}{\textbf{Models}}\\
			\cmidrule(r{0.02in}){4-6}
			&&&\textbf{Base} & \textbf{Large} & \textbf{Huge}\\
			\midrule
			
			\multirow{3}{*}{\textbf{CamCAN}} & \multirow{3}{*}{MAE $\downarrow$}
			& fMRI-only  
			& 12.09{\fontsize{5pt}{6pt}\selectfont$\pm$0.78}
			& 11.71{\fontsize{5pt}{6pt}\selectfont$\pm$0.67}
			& 9.51{\fontsize{5pt}{6pt}\selectfont$\pm$0.51}\\
			
			&& MEG-only      
			& 8.99{\fontsize{5pt}{6pt}\selectfont$\pm$0.14}  
			& 8.86{\fontsize{5pt}{6pt}\selectfont$\pm$0.21}
			& \textbf{7.87{\fontsize{5pt}{6pt}\selectfont$\pm$0.22}}  \\
			\rowcolor{gray!10}
			&& \textbf{Multimodal Fusion} 
			& \textbf{8.87{\fontsize{5pt}{6pt}\selectfont$\pm$0.09}}
			& \textbf{8.78{\fontsize{5pt}{6pt}\selectfont$\pm$0.07}}
			& 8.21{\fontsize{5pt}{6pt}\selectfont$\pm$0.11} \\
			\midrule
			\multirow{3}{*}{\textbf{LEMON}} & \multirow{3}{*}{BAC(\%) $\uparrow$}
			& fMRI-only  
			& 58.95{\fontsize{5pt}{6pt}\selectfont$\pm$2.13}
			& 62.85{\fontsize{5pt}{6pt}\selectfont$\pm$4.29}
			& 61.79{\fontsize{5pt}{6pt}\selectfont$\pm$1.29} \\
			
			&& EEG-only      
			& 76.05{\fontsize{5pt}{6pt}\selectfont$\pm$0.44}
			& 78.04{\fontsize{5pt}{6pt}\selectfont$\pm$1.59}
			& 80.39{\fontsize{5pt}{6pt}\selectfont$\pm$0.59} \\
			\rowcolor{gray!10}
			&& \textbf{Multimodal Fusion} 
			& \textbf{76.77{\fontsize{5pt}{6pt}\selectfont$\pm$0.60}}
			& \textbf{78.32{\fontsize{5pt}{6pt}\selectfont$\pm$1.64}}
			&\textbf{81.10{\fontsize{5pt}{6pt}\selectfont$\pm$0.54}}\\
			\bottomrule
		\end{tabular}
	\end{small}
	\vskip -0.1in
\end{table*}

\vspace{-0.5em}
\subsection{Performance Gains via Multimodal Fusion}
\vskip -0.1in
Beyond unimodal evaluation, we assess multimodal fusion on two paired-modality benchmarks: brain-age prediction on CamCAN (fMRI + MEG) and age-group classification on LEMON (fMRI + EEG). Fusion is implemented via serial resampling with ARNESS, which integrates heterogeneous modalities into a unified fixed-length sequence without additional fusion modules or auxiliary alignment losses. Regarding sequence construction, we adopt an empirical ordering strategy in which the dominant modality (i.e., achieving higher unimodal performance) is placed at the end of the sequence to maximize its influence on the final representation.
As shown in Table~\ref{multimodal}, multimodal fusion consistently outperforms unimodal baselines across nearly all model scales. For both tasks, electrophysiological signals (MEG/EEG) achieve stronger unimodal performance than fMRI, reflecting the temporal sensitivity of these benchmarks. Importantly, incorporating fMRI further improves results by injecting complementary information. For example, with Brain-OF Large, fusion reduces the CamCAN MAE to 8.78 (vs. 8.86 with MEG-only) and increases LEMON BAC to 78.32\% (vs. 78.04\% with EEG-only). 
These results demonstrate that multimodal fusion in Brain-OF yields improved predictive performance compared to unimodal settings.\vspace{-1.0em}

\subsection{Visualization of Model Explanations}
\begin{figure*}[ht]
	\vskip -0.1in
	\centering
	\includegraphics[width=0.99\linewidth]{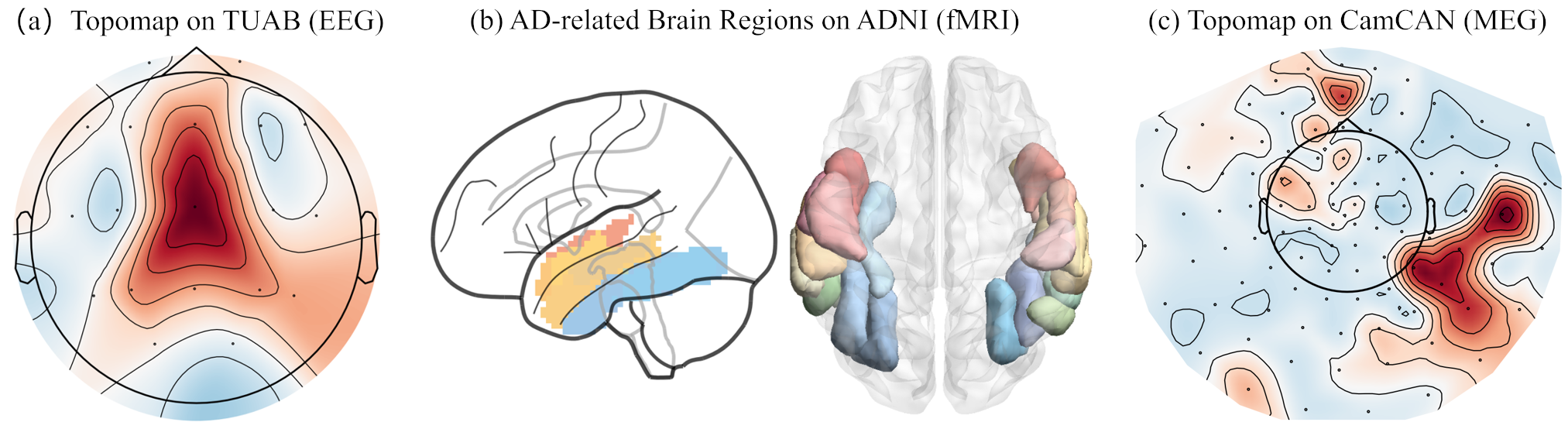}
	\vskip -0.05in
	\caption{
		\textbf{Visualization of Brain-OF Interpretability.}
	}
	\label{explanation}
	\vskip -0.15in
\end{figure*}

\vskip -0.1in
To interpret the representations learned by Brain-OF Base, we employ the occlusion method \cite{zeiler2014visualizing} to quantify the contribution of each channel/ROI to the model's predictive performance across three representative downstream tasks.
For TUAB EEG abnormality detection (Figure~\ref{explanation}(a)), the dataset contains heterogeneous pathological EEG patterns, including focal and generalized abnormalities, and therefore no single scalp topography is expected to characterize all abnormal recordings. Brain-OF assigns relatively high importance to frontal and central electrodes. This pattern is compatible with generalized or rhythmic slowing, which can be bilateral and anteriorly prominent \cite{britton2016electroencephalography}. Previous work on TUAB further showed that pathological EEG classification relies strongly on low-frequency delta and theta activity, particularly from temporal channels, while clinical reports also contain descriptions of both temporal and rhythmic frontal slowing~\cite{schirrmeister2017deep}.
For Alzheimer’s disease (AD) classification on ADNI (Figure~\ref{explanation}(b)), the top 10\% most influential ROIs are primarily located in the fusiform gyrus, as well as the middle and inferior temporal gyri. These regions are relevant to visual and memory-related processing and are among the temporal neocortical regions reported to exhibit atrophy associated with progression toward Alzheimer’s disease~\cite{vockert2024cognitive, convit2000atrophy}. Thus, the visualization suggests that Brain-OF uses disease-relevant temporal patterns for classification and we regard this correspondence as evidence of biological plausibility.
Finally, the sensor-level contribution map for MEG-based age prediction on CamCAN (Figure~\ref{explanation}(c)) shows that the most influential sensors are primarily located over temporal and parietal scalp areas. Prior MEG studies have shown that age can be predicted from electrophysiological dynamics that temporal and visual cortical signals contain particularly informative age-related features~\cite{stier2025temporal, lee2022deep}. The observed sensor distribution is therefore consistent with known age-related changes, further supporting the neurobiological plausibility for MEG.
Taken together, these visualizations provide a complementary plausibility check, examining whether Brain-OF makes its predictions on neurobiologically meaningful representations across modalities rather than on obvious artifacts or dataset-specific shortcuts.

\section{Conclusion}
\vskip -0.1in
In this work, we introduce \textbf{Brain-OF}, an omnifunctional brain foundation model designed to harness the complementary spatiotemporal characteristics of fMRI, EEG and MEG. To address the heterogeneity of signal resolutions across modalities, we propose the \textit{Any-Resolution Neural Signal Sampler}, which projects diverse inputs into a shared semantic space. Our architecture further addresses semantic complexity through a backbone integrating \textit{DINT attention} and \textit{Sparse Mixture of Experts}, where shared and routed experts capture modality-invariant knowledge and modality-specific representations, respectively. Crucially, to enforce deep understanding of brain dynamics, we propose \textit{Masked Temporal-Frequency Modeling}, a pretraining objective that compels joint reconstruction of signals in both time and frequency domains.
Extensive experiments across unimodal and multimodal tasks demonstrate that Brain-OF achieves superior performance, highlighting the benefits of large-scale joint pretraining across heterogeneous brain signals. We hope that Brain-OF paves the way toward general-purpose, scalable and interpretable brain foundation models for both unimodal and cross-modal neuroscience applications.

\begin{ack}
The authors gratefully acknowledge the Gauss Centre for Supercomputing e.V. (www.gauss-centre.eu) for funding this project by providing computing time through the John von Neumann Institute for Computing (NIC) on the GCS Supercomputer JUWELS\cite{JUWELS} at Jülich Supercomputing Centre (JSC). Data were provided (in part) by the Donders Institute for Brain, Cognition and Behaviour at Radboud University (Nijmegen, The Netherlands) and the Cambridge Centre for Ageing and Neuroscience (CamCAN).
\end{ack}

\bibliographystyle{unsrtnat}
\bibliography{references}
%
%
%
%
%
%
%
%
%
%

\newpage
\appendix

\section{Details of Datasets}
\label{detailsofDatasets}
\vskip -0.1in
In this study, we integrate three heterogeneous functional brain imaging modalities (i.e., fMRI, EEG and MEG) to develop the omnifunctional brain foundation model (Brain-OF). These modalities differ substantially in their temporal, spatial, and spectral resolutions, giving rise to both semantic richness and considerable heterogeneity. Within each modality, further variability arises from heterogeneous acquisition protocols and scanners across international sites, reflected in diverse sampling rates (MEG/EEG), repetition times and field strengths (3 T vs. 7 T for fMRI), number of channels/sensors, recording durations and experimental paradigms (resting vs. task state).

\begin{table*}[h]
	\vskip -0.15in
	\caption{Summary of all pretraining datasets, covering 3 heterogeneous functional neuroimaging modalities (fMRI, EEG, MEG) across 37 public datasets, encompassing over 32K unique participants, approximately 5.9M samples, and 2.8 TB of preprocessed data.}
	\label{pretraining_datasets}
	\centering
	\setlength{\tabcolsep}{3pt}
	\begin{small}
			\begin{tabular}{l@{\hspace{2pt}}ccccccr}
				\toprule
				Modalities  & Datasets & Paradigms & SR/TR & \#Chs/ROIs & Durations & \#Participants & \#Samples  \\
				\midrule
				\multirow{13}{*}{EEG} 
				& FACED & task & 250Hz & 32 & 10s & 123 & 10,332 \\
				& PhysioNet-MI & task & 160Hz & 64 & 4s & 109 & 18,697 \\
				& PREDICT & rest & 5,000Hz & 63 & 10s & 159 & 17,511 \\
				& REEG-BACA & rest & 1,000Hz & 64 & 10s & 608 & 53,089 \\
				& SEED & task & 1,000Hz & 62 & 10s & 15 & 14,895 \\
				& SEED-DV & task & 1,000Hz & 62 & 10s & 20 & 7,644 \\
				& SEED-SD & task & 1,000Hz & 62 & 10s & 40 & 38,120 \\
				& SEED-VIG & task & 200Hz & 17 & 8s & 21 & 20,332 \\
				& SEED-VII & task & 1,000Hz & 62 & 10s & 20 & 26,257 \\
				& SEED-VLA/VRW & task & 300Hz & 18 & 10s & 20 & 17,287 \\
				& TDBRAIN & rest & 500Hz & 26 & 10s & 1,274 & 26,422 \\
				& TUEG & rest & 250Hz & 19 & 10s & 13,928 & 4,562,151  \\
				& WBCIC-MI & task & 1,000Hz & 58 & 4s & 62 & 40,290  \\
				\midrule
				\multirow{18}{*}{fMRI}
				& ABIDE-I  & rest & multi-site & 246 & - & 1,112 & 1,152 \\
				& ABIDE-II  & rest & multi-site & 246 & - & 1,114 & 1,614 \\
				& ABRIM  & rest \& task & 1,000ms & 246 & 320s & 282 & 562 \\
				& AOMIC-ID1000  & task & 2,200ms & 246 & 660s & 881 & 881 \\
				& AOMIC-PIOP1  & rest \& task & 750ms & 246 & ~360s & 216 & 1,236 \\
				& AOMIC-PIOP2  & rest \& task & 2,000ms & 246 & ~480s & 226 & 896 \\
				& CHCP  & rest \& task & 710ms & 246 & 227s & 366 & 3,379 \\
				& CoRR  & rest & multi-site & 246 & - & 1,629 & 5,212 \\
				& DLBS  & rest \& task & 2,000ms & 246 & 374s & 464 & 6,839 \\
				& NADR & rest & 3,000ms & 246 & 606s & 301 & 596 \\
				& Narratives & task & 1,500ms & 246 & 480s & 345 & 1,321  \\
				& NKI-RS & rest & 1,400ms & 246 & 448s & 1,344 & 16,408  \\
				& Paingen placebo  & task & 460ms & 246 & 132s & 395 & 3,143 \\
				& QTIM  & rest \& task & 2,100ms & 246 & 315s & 1,202 & 2,453 \\
				& REST-meta-MDD  & rest & multi-site & 116 & - & 2,380 & 2,380 \\
				& SALD  & rest & 2,000ms & 246 & 480s & 493 & 493 \\
				& SpaceTop  & task & 460ms & 246 & 200s & 101 & 11,660 \\
				& UCLA CNP  & rest \& task & 2,000ms & 246 & 304s & 272 & 1,969 \\
				\midrule
				\multirow{11}{*}{\begin{tabular}{l}MEG \\/\\Multimodal \end{tabular}}
				& CamCAN-MEG  & rest \& task & 1,000Hz & 306 & 2s & 456 & 279,965  \\
				& CamCAN-fMRI  & rest \& task & 1,970ms & 246 & 520s & 465 & 1,763  \\
				& HCP-YA-MEG  & rest \& task & 2,000Hz & 248 & 2s & 89 & 241,115  \\
				& HCP-YA-fMRI  & rest \& task & 720ms & 246 & 230s & 1,113 & 32,281  \\
				& LEMON-EEG  & rest & 2,500Hz & 61 & 10s & 144 & 12,235 \\
				& LEMON-fMRI  & rest & 1,400ms & 246 & 448s & 160 & 320 \\
				& MOUS-MEG  & rest \& task & 1,200Hz & 275 & 2s & 204 & 66,322 \\
				& MOUS-fMRI  & rest \& task & 1,680ms & 246 & 420s & 204 & 678 \\
				& OMEGA  & rest & 2,400Hz & 275 & 2s & 644 & 164,834  \\
				& WAND-MEG  & rest \& task & 1,200Hz & 275 & 2s & 170 & 155,474 \\
				& WAND-fMRI  & rest \& task & 2,000ms & 246 & 638s & 170 & 776 \\
				\midrule
				\textbf{Total} & \textbf{fMRI, EEG, MEG} & \textbf{rest \& task} & - & - & - & \textbf{32,278} & \textbf{5,870,984} \\
				\bottomrule
			\end{tabular}
	\end{small}
	\begin{minipage}{\linewidth}
		\footnotesize
		Some datasets include multiple sampling rates (SR) or repetition times (TR) due to differences in acquisition protocols across different phases or paradigms (e.g., resting-state vs. task-based scans). For conciseness, we report the primary SR/TR for each dataset in the “SR / TR” column. Similarly, the reported durations primarily correspond to resting-state recordings. For detailed information, please refer to the original publications. Chs: channels. ROIs: regions of interest.
	\end{minipage}
	\vskip -0.05in
\end{table*}
\vskip -0.05in

This diversity further supports the construction of a pluralistic pretraining corpus spanning multiracial, multicultural populations, including both healthy individuals and clinical patients. In total, our pretraining dataset aggregates 37 publicly available datasets across fMRI, EEG and MEG, comprising 32,278 unique participants, 5,870,984 samples, and 2.8 TB of preprocessed data. Detailed statistics for each dataset can be found in Table~\ref{pretraining_datasets}.

\vskip -0.05in
\subsection{Data Preprocessing} 
\vskip -0.05in
If the public datasets provide official preprocessed data, we use them directly. Otherwise, we apply minimal, modality-specific preprocessing pipelines to remove artifacts. Raw EEG and MEG signals are preprocessed using MNE-Python \cite{gramfort2014mne}, and fMRI data are preprocessed with fMRIPrep \cite{esteban2019fmriprep}. After preprocessing, z-score normalization is applied to all signals to harmonize their value ranges across fMRI, EEG and MEG.

\textbf{EEG.} To enhance stationarity, the initial and final 10 seconds of EEG signals are discarded. Signals are then band-pass filtered (1–80 Hz, zero-phase Hamming-windowed sinc FIR, –6 dB cutoff) and notched at the local power-line frequency (50 or 60 Hz). Filtered signals are resampled to 200 Hz, segmented into non-overlapping 10-second samples for resting-state data, and re-referenced to the common average.

\textbf{MEG.} MEG preprocessing similarly undergoes band-pass filtering (1–80 Hz, zero-phase Hamming-windowed sinc FIR, –6 dB cutoff) and notching at the power-line frequency. Filtered signals are resampled to 200 Hz and segmented into non-overlapping 2-second samples for resting-state recordings.

\textbf{fMRI.} For each BOLD series, a reference volume and its skull-stripped version are estimated. Head-motion parameters (volume-to-reference transforms and corresponding rotations/translations) are computed, followed by slice-timing correction using 3dTshift (AFNI). Functional volumes are co-registered to the T1-weighted anatomical reference via boundary-based registration (9 degrees of freedom) to minimize distortions. Motion correction, susceptibility distortion correction, BOLD-to-T1w, and T1w-to-template (MNI152NLin2009cAsym) warps are concatenated and applied in one interpolation step using antsApplyTransforms (ANTs). Nuisance regressors, including global signal, mean signals from white matter and CSF, tCompCor, aCompCor, framewise displacement, six motion parameters, DVARS, and spike regressors, are regressed out. Residual signals are spatially smoothed (6 mm FWHM Gaussian kernel) and subjected to Wishart filtering. Finally, regional time series are extracted from 246 parcels using the Brainnetome atlas \cite{fan2016human}.

\vskip -0.1in
\subsection{Pretraining Datasets}
\begin{itemize}
	\item \textbf{FACED} \cite{chen2023large} is a large finer-grained affective computing EEG dataset comprising 32-channel recordings at sampling rate of 250 or 1,000 Hz from 123 subjects (ages 17-38). Each participant was instructed to watch 28 emotion-elicitation video clips spanning nine discrete emotional states: amusement, inspiration, joy, tenderness, anger, fear, disgust, sadness and neutral emotion.
	\item \textbf{PhysioNet-MI} \cite{schalk2004bci2000} is a large-scale motor imagery EEG collections, comprising 64-channel recordings (160 Hz sampling rate) from 109 healthy participants performing left fist, right fist, both fists, and both
	feet imagery. For pretraining, we use 4-second trials after preprocessing.
	\item \textbf{PREDICT} \cite{ds005486:1.0.1,chowdhury2025predicting} is a longitudinal pain-related EEG dataset comprising 5-minute, eyes-closed, 63-channel resting-state recordings from 159 participants (ages 18–44) with no history of chronic pain, neurological disorders, or psychiatric conditions. EEG data were collected on Days 0, 2, and 5 of the experimental protocol, providing multiple sessions per participant for identifying individual pain sensitivity in healthy adults.
	\item \textbf{REEG-BACA} \cite{getzmann2024resting} is a large-scale longitudinal resting-state EEG dataset collected from 608 participants (ages 20-70), with 208 participants returning for a follow-up session approximately 5 years later. High-density rs-EEG was acquired using a 64-channel elastic cap (10–20 system) at 1,000 Hz sampling rate. Recordings were obtained both before and after a 2-hour battery of cognitive tasks.
	\item \textbf{SEED Series} is a family of EEG datasets designed for affective computing and related cognitive neuroscience research. It includes \textbf{SEED} \cite{zheng2015investigating}, \textbf{SEED-VIG} \cite{1741-2552-14-2-026017}, \textbf{SEED-VII} \cite{10731546}, \textbf{SEED-VLA/VRW} \cite{Luo_2024}, \textbf{SEED-DV} \cite{liu2024eegvideo} and \textbf{SEED-SD} \cite{seedsd}. All datasets were recorded using a 62-channel ESI NeuroScan system at a 1,000 Hz sampling rate during emotion-related paradigms spanning positive, negative and neutral valence, as well as more fine-grained affective states (e.g., amusement, fear, sadness). In total, the SEED Series provides EEG recordings from 136 unique participants, predominantly young adults.
	\item \textbf{TDBRAIN} \cite{van2022two} is a large single-site lifespan EEG database (ages 5–89), comprising resting-state recordings from 1,274 psychiatric patients. The psychophysiological data include 26 channel EEG acquired at a sampling rate of 500 Hz.
	\item \textbf{TUEG} \cite{obeid2016temple} is the largest publicly available collection of clinical EEG dataset, comprising 69,652 recordings from 14,987 subjects across 26,846 sessions collected at Temple University Hospital. The dataset exhibits substantial variability in channel configurations, sampling rates and recording durations, reflecting the heterogeneity of real-world clinical practice. To prevent information leakage and ensure reliable downstream evaluation, we exclude all subjects that appear in the TUAB test set.
	\item \textbf{WBCIC-MI} \cite{yang2025multi} is a high-quality motor imagery EEG dataset comprising recordings from 62 healthy participants (ages 17–30) across three sessions. For pretraining, we incorporate the officially released preprocessed signals: 4-second trials recorded with 58 EEG channels.
	
	\item \textbf{ABIDE} is an international collaborative initiative that aggregates resting-state fMRI, structural MRI and phenotypic data collected independently across more than 24 imaging centers worldwide. It contains two large-scale collections: \textbf{ABIDE-I} \cite{di2014autism}, which includes 1,112 participants (ages 7–64; 539 individuals with autism spectrum disorder and 573 controls), and \textbf{ABIDE-II} \cite{di2017enhancing}, which includes 1,114 participants (521 individuals with autism spectrum disorder and 593 controls).
	\item \textbf{ABRIM} \cite{null_null_2023_the_advanced_brain_i} is a cross-sectional adult lifespan MRI collection comprising structural and functional MRI scans from 295 healthy participants (ages 18–80), along with cognitive assessments, proxy measures of cognitive reserve, and 7-day actigraphy recordings of sleep-wake rhythms. After removing corrupted and missing data, a total of 282 subjects are included in our pretraining dataset.
	\item \textbf{AOMIC} \cite{snoek2021amsterdam} is one of the largest and richest open-access 3 T neuroimaging resources focused on young adults. It comprises three complementary datasets: \textbf{ID1000} (N = 928, ages 19–26), \textbf{PIOP1} (N = 216, age 18–26) and \textbf{PIOP2} (N = 226, ages 18–26), together providing 1,370 unique participants. Each sub-cohort includes high-resolution T1-weighted structural MRI, diffusion-weighted imaging, multiple resting-state runs, and an exceptionally broad battery of task-based fMRI paradigms (naturalistic movie watching, emotion processing, working memory, face recognition, cognitive conflict and response inhibition).
	\item \textbf{CHCP} \cite{ge2023increasing} is the Chinese Human Connectome Project, comprising neuroimaging and behavioral data from 366 healthy adults living in an Eastern culture. Its acquisition protocols closely mirror the original HCP-YA, including high-resolution T1w/T2w structural MRI, resting-state and task-based fMRI (seven tasks), and high-angular-resolution diffusion MRI, as well as behavioral data based on Chinese population.
	\item \textbf{CoRR} \cite{zuo2014open} is the largest publicly available test-retest resting-state fMRI consortium, aggregating data from 36 independent research groups across 18 international sites. The dataset comprises 5,093 resting-state fMRI scans from 1,629 unique participants (ages 6–88), with many individuals providing multiple sessions. This multi-site, multi-scanner design introduces natural acquisition variability highly beneficial for learning robust representations.
	\item \textbf{DLBS} \cite{park2025dallas} is a longitudinal adult lifespan neuroimaging dataset designed to characterize brain–cognition relationships across aging. A total of 464 participants (ages 21–89 at baseline) completed the first assessment, with 338 and 224 participants returning for second and third assessment epochs, respectively, at intervals of approximately 3.5–5 years. Each epoch includes an extensive neuropsychological battery, questionnaires assessing physical health, psychosocial factors and brain health, as well as multimodal MRI acquisitions: T1-weighted structural imaging, diffusion-weighted imaging, hypercapnia imaging, arterial spin labeling and four functional MRI scans (resting-state and task-based fMRI), totaling 6,893 fMRI sessions across all visits.
	\item \textbf{NADR} \cite{spreng2022neurocognitive} is a multisite neuroimaging dataset designed to characterize age-related differences in brain function. It includes 301 healthy adults, comprising 181 young adults (ages 18–34) and 120 older adults (ages 60–89). Each participant underwent two 10-minute multi-echo resting-state fMRI runs and a high-resolution T1-weighted MPRAGE structural scan acquired on 3T MRI scanners across two sites.
	\item \textbf{Narratives} \cite{nastase2021narratives} is a naturalistic paradigm dataset comprising 891 fMRI runs from 345 healthy adults (age 18–53 years) who listened to 27 different spoken stories. These stories vary widely in length, speaker, and content, together providing ~4.6 hours of unique auditory narrative stimulation (~43,000 words).
	\item \textbf{NKI-RS} \cite{tobe2022longitudinal} is a large-scale longitudinal lifespan neuroimaging initiative comprising 1,344 participants, designed to characterize developmental and aging trajectories across the human lifespan. The dataset provides comprehensive psychiatric, medical, behavioral and cognitive phenotyping, along with multimodal brain imaging across multiple follow-up visits (resting-state fMRI, diffusion MRI, structural morphometric MRI and arterial spin labeling), genetic data and actigraphy measures.
	\item \textbf{Paingen placebo} \cite{botvinik2024placebo} is a large-scale, high-quality pain and placebo analgesia fMRI collection comprising 395 healthy adults (age 30–43 years). Each participant completed four fMRI runs during a well-controlled thermal pain and placebo manipulation paradigm, plus anatomical scans and field maps.
	\item \textbf{QTIM} \cite{blokland2011heritability,sinclair2015heritability} is one of the largest twin neuroimaging cohorts worldwide. It comprises resting-state and task-based fMRI from 1,202 young adult twins and siblings (ages 12–30 at baseline), including a test-retest subsample (N = 78, ~3.5-month interval) and longitudinal adolescent twin cohorts scanned at ages 12 and 16 (N = 88 each), with a subset rescanned four years later (N = 24 at both 12 and 16; N = 38 at 16 and 20).
	\item \textbf{REST-meta-MDD} \cite{chen2022direct} is a large-scale resting-state fMRI consortium for major depressive disorder (MDD), aggregating data from 25 research groups across 17 hospitals in China. It comprises 2,428 resting-state fMRI scans from 1,300 patients with MDD and 1,128 healthy controls, providing a diverse and clinically representative sample for studying MDD alterations in brain function.
	\item \textbf{SALD} \cite{wei2018structural} is a cross-sectional adult lifespan neuroimaging dataset comprising structural MRI and resting-state fMRI from 493 healthy adults (ages 19–80). All scans were acquired on a 3 T Siemens Trio scanner at the Brain Imaging Center of Southwest University, China.		
	\item \textbf{SpaceTop} \cite{jung2025spacetop}	is a landmark naturalistic and multidimensional neuroimaging dataset comprising 101 healthy young adults, each completing approximately 40 fMRI runs. It uniquely integrates a broad spectrum of experimental paradigms, including cognitive, affective, social and somatic/interoceptive tasks, together with structural T1-weighted imaging, multi-shell diffusion MRI and autonomic physiological recordings.
	\item \textbf{UCLA CNP} \cite{poldrack2016phenome} is a publicly shared neuroimaging dataset comprising 272 adults (ages 21–50): 130 healthy controls and 142 patients (50 schizophrenia, 49 bipolar disorder, 43 ADHD). The dataset provides an extensive set of imaging modalities, including task-based fMRI across multiple cognitive domains, resting-state fMRI, structural MRI and high–angular-resolution diffusion MRI.
	\item \textbf{CamCAN} \cite{taylor2017cambridge} is a large multimodal, cross-sectional adult lifespan dataset from 656 healthy adults (ages 18–87). It includes 3T resting-state and task fMRI (movie-watching and sensorimotor task) as well as high-density MEG (306-channel Elekta Neuromag, 1,000 Hz sampling rate) during both resting state and the same sensorimotor task. For pretraining, we incorporate all available fMRI and MEG data except subjects assigned to downstream validation and test sets. This strict subject-level exclusion prevents contamination while maximizing pretraining data volume.
	\item \textbf{HCP Young Adult} \cite{van2013wu} includes multiple 3T and 7T fMRI runs as well as MEG recordings from 1,113 young healthy adult (ages 22-35). For pretraining, we incorporate all available functional modalities: 3T resting-state and task fMRI, 7T resting-state and task fMRI and both resting-state and task MEG.
	\item \textbf{LEMON} \cite{babayan2019mind} is a multimodal neuroimaging dataset that includes both fMRI and EEG recordings. Participants underwent a 3T fMRI scan and a 61-channel resting-state EEG acquisition. Due to the scarcity of high-quality multimodal datasets, we also incorporate LEMON as a downstream evaluation task. To avoid contamination in downstream evaluation, subjects overlapping with the LEMON validation and test splits are excluded from the pretraining dataset.
	\item \textbf{MOUS} \cite{schoffelen2019204} is a multimodal neuroimaging dataset collected from 204 healthy adults (ages 18-33). The protocol includes whole-brain fMRI and high-density 275-channel MEG acquired during a Dutch language paradigm as well as during resting state. For pretraining, we incorporate all available MEG and fMRI sessions from the MOUS cohort.
	\item \textbf{OMEGA} \cite{niso2016omega} is a large, open-access collection of resting-state MEG recordings led by the McConnell Brain Imaging Centre (BIC) at the Montreal Neurological Institute (MNI), McGill University, in collaboration with the Université de Montréal. It is a continuously expanding resource; for this study, we include resting-state MEG from 644 participants (444 healthy controls and 200 patients with various psychiatric conditions).
	\item \textbf{WAND} \cite{mcnabb2025wand} is a multiscale, multimodal neuroimaging dataset comprising in vivo brain data from 170 healthy adults (ages 18–63), including 3T and 7T structural and functional MRI as well as 275-channel MEG recorded at a 1,200 Hz sampling rate. We incorporate all available resting- and task-state fMRI (3 T and 7 T) and MEG data into the pretraining dataset.
\end{itemize}

\subsection{Downstream Task Datasets}
\label{downstreamTaskDatasets}
\begin{itemize}
	\item \textbf{SEED-V} \cite{liu2021comparing} is an emotion recognition EEG benchmark collected from 20 participants using a 62-channel ESI NeuroScan system at a 1,000 Hz sampling rate. It includes five emotion categories (happy, sad, fear, disgust and neutral) recorded across three sessions. Each session contains 15 trials, yielding a total of 117,744 1-second samples. Following the conventional strategy, we divide the fifteen trials of each session into three equal parts (5:5:5) as the training, validation and test sets, respectively.
	\item \textbf{CHB-MIT} \cite{shoeb2009application, goldberger2000physiobank} is an intractable-seizure EEG database comprising recordings from 22 pediatric subjects (ages 1.5–22), acquired at 256 Hz with the international 10–20 system at Children’s Hospital Boston. For our experiments, all recordings are resampled to 200 Hz and	segmented into 326,993 10-second samples. Notably, seizure events constitute only 0.68\% of all samples, resulting in a severe class imbalance. We follow the conventional subject-wise split: subjects 1 to 19 for training, subjects 20, 21 for validation, and subjects 22,23 for testing.
	\item \textbf{TUAB} \cite{obeid2016temple} is a subset of the TUEG corpus, containing EEG recordings from 1,385 healthy controls (normal) and 998 subjects with a brain disorder (abnormal), totaling 2,993 sessions. After preprocessing, we obtain 409,455 10-second samples for binary classification task. The official evaluation split is used as our test set. To avoid potential data leakage effects, all subjects included in the TUAB test set are excluded from the TUEG dataset.
	\item \textbf{ADNI} \cite{jack2008alzheimer} is a longitudinal, multisite neuroimaging initiative that includes structural, functional and molecular imaging, along with biofluid biomarkers, cognitive assessments, genetic data and demographic information from healthy older adults, individuals with mild cognitive impairment and patients with Alzheimer’s disease (AD). For this downstream task, we incorporate resting-state fMRI data from 222 participants (110 patients with AD and 112 age- and gender-matched healthy controls), yielding a total of 508 scanning sessions. After preprocessing, a total of 618 samples are used.
	\item \textbf{ADHD-200} \cite{adhd2012adhd} is a large-scale, multi-site benchmark for neurodevelopmental classification, comprising resting-state fMRI, anatomical MRI and phenotypic data from 926 children and adolescents (573 typically developing controls, 353 individuals with ADHD) across eight international sites. In total, participants contributed 1,369 resting-state fMRI sessions. For the downstream binary ADHD vs. control classification task, all 1,369 sessions are used after mini-preprocessing.
	\item \textbf{CamCAN} \cite{taylor2017cambridge} is a widely used benchmark for studying age-related changes in brain function, particularly in MEG. It contains both MEG and fMRI recordings from 656 healthy adults spanning the adult lifespan (ages 18–87). We use this dataset for chronological brain-age prediction under three evaluation settings: (i) unimodal resting-state fMRI (652 8min40s recordings), (ii) unimodal resting-state MEG (non-overlapping 177,044 2 second segments), and (iii) multimodal fusion of fMRI and MEG. To eliminate any risk of information leakage, all subjects included in the CamCAN validation and test splits are strictly excluded from the pretraining dataset.
	\item \textbf{LEMON} \cite{babayan2019mind} consists of 227 healthy participants divided into a young cohort (N = 153, age 25.1 $\pm$ 3.1 years, range 20–35 years, 45 female) and an elderly cohort (N = 74, age 67.6 $\pm$ 4.7 years, range 59–77 years, 37 female). For downstream age-group classification (young vs. elderly), we evaluate on three settings: (i) unimodal fMRI using 438 7.5-minute resting-state scans, (ii) unimodal EEG using non-overlapping 17,235 10-second segments, and (iii) multimodal fusion of fMRI and EEG. To prevent information leakage and ensure reliable downstream evaluation, all subjects in the LEMON validation and test splits are excluded from the pretraining dataset.
	\item \textbf{HAD} \cite{zhou2023large} is a task-based fMRI dataset for human action recognition, consisting of fMRI responses to 21,600 video clips from 30 participants. The clips encompass 180 human action categories, covering a wide range of complex, real-world activities.	For the downstream task, we perform a highly challenging 180-way event-related decoding classification of the evoked fMRI recordings. We employ a strict subject-wise split: subjects 1–24 for training (17,280 samples), 25–27 for validation (2,160 samples), and 28–30 for testing (2,160 samples).
	\item \textbf{BOLD5000} \cite{chang2019bold5000} is a slow event-related fMRI dataset collected from four participants across 16 sessions. During scanning, participants viewed nearly 5,000 distinct static images representing real-world scenes from the SUN, COCO, and ImageNet datasets. For the downstream evaluation, we conduct a 3-way cognitive decoding task to classify the source domain of each visual stimulus (COCO, ImageNet, or SUN). We adopt a strict subject-wise split: subjects 1 and 2 for training (10,508 samples), subject 3 for validation (5,254 samples), and subject 4 for testing (3,108 samples).
\end{itemize}

To explicitly characterize generalization to unseen data, sites, and subjects, as well as the partial overlap between pretraining and downstream datasets, we summarize the relationship between the pretraining corpus and each downstream dataset in Table~\ref{tab:dataset_overlap}. Our evaluation covers two major settings: \emph{dataset-level generalization}, where the downstream dataset is entirely absent from pretraining, and \emph{subject-level generalization}, where the same dataset or a related parent corpus contributes training subjects to pretraining, while all downstream validation and test subjects are strictly excluded.

\begin{table*}[h]
	\vskip -0.05in
	\centering
	\caption{Relationship between the pretraining corpus and downstream evaluation datasets. 
		$\checkmark$ denotes inclusion in pretraining, $\times$ denotes no inclusion or overlap, and $\triangle$ denotes partial overlap through the same dataset or a related parent corpus. 
		PT denotes pretraining. All downstream validation and test subjects are strictly excluded from pretraining.}
	\label{tab:dataset_overlap}
	\small
	\begin{small}
	\setlength{\tabcolsep}{3pt}
	\begin{tabular}{lccccl}
		\toprule
		\textbf{Dataset} &
		\textbf{Pretraining} &
		\textbf{Relation} &
		\textbf{Train Subjects in PT} &
		\textbf{Val./Test Overlap} &
		\textbf{OOD Level} \\
		\midrule
		SEED-V     
		& $\times$    & Independent & $\times$    & $\times$ 
		& Dataset + Subject \\
		
		CHB-MIT    
		& $\times$    & Independent & $\times$    & $\times$ 
		& Dataset + Site + Subject \\
		
		TUAB        
		& $\triangle$ & Partial     & $\checkmark$ & $\times$ 
		& Subject \\
		
		ADNI        
		& $\times$    & Independent & $\times$    & $\times$ 
		& Dataset + Site + Subject \\
		
		ADHD-200    
		& $\times$    & Independent & $\times$    & $\times$ 
		& Dataset + Site + Subject \\
		
		CamCAN      
		& $\triangle$ & Partial     & $\checkmark$ & $\times$ 
		& Subject \\
		
		LEMON       
		& $\triangle$ & Partial     & $\checkmark$ & $\times$ 
		& Subject \\
		
		HAD         
		& $\times$    & Independent & $\times$    & $\times$ 
		& Dataset + Site + Subject \\
		
		BOLD5000    
		& $\times$    & Independent & $\times$    & $\times$ 
		& Dataset + Site + Subject \\
		\bottomrule
	\end{tabular}
\end{small}
\end{table*}

\section{Baseline Models}
\label{baselines}
To compare Brain-OF with representative baselines, we include both task-specific models and recent foundation models across different modalities, as summarized in Table~\ref{baseline_model}.

\begin{table*}[t]
	\vskip -0.05in
	\caption{\textbf{Overview of baseline models.} We categorize baselines into task-specific models (trained from scratch) and foundation models (pretrained). \textbf{Params} denotes parameter count in millions (M); \textbf{GFLOPs} measures computational cost per inference sample.}
	\label{baseline_model}
	\centering
	\begin{small}
		\begin{tabular}{lcccc}
			\toprule
			\textbf{Models} & \textbf{Params (M)} & \textbf{GFLOPs} & \textbf{Architecture} & \textbf{Pretraining Modality} \\
			\midrule
			\multicolumn{5}{l}{\textit{Task-Specific Models (Trained from scratch)}} \\
			\midrule
			EEGNet & 0.004 & 0.13 & CNN & \ding{55} \\
			EEGConformer & 0.81  & 0.58 & Convolutional Transformer & \ding{55} \\
			ContraWR & 1.57 & 0.28 & CNN & \ding{55} \\
			CNN-Transformer & 3.16 & 0.33 & CNN + Transformer & \ding{55} \\
			FFCL & 2.47 & 0.96 & CNN + LSTM & \ding{55} \\
			ST-Transformer & 3.43 & 0.22 & Transformer & \ding{55} \\
			BrainNetTF & 5.56 & 0.74 & Graph-structured Transformer & \ding{55} \\
			BrainNetCNN & 1.13 & 0.51 & Connectivity-based CNN & \ding{55} \\
			\midrule
			\multicolumn{5}{l}{\textit{Foundation Models (Pretrained)}} \\
			\midrule
			LaBraM Base & 5.82 & 2.70 & Transformer & EEG \\
			CBraMod & 4.88 & 1.86 & Criss-Cross Transformer & EEG \\
			BrainHarmonix-F & 85.25 & 1450.00 & Vision Transformer & fMRI \\
			Brain-JEPA & 85.08 & 1271.79 & Vision Transformer & fMRI\\
			BrainOmni Base & 32.71 & 11.43 & Transformer & EEG + MEG\\
			\bottomrule
		\end{tabular}
	\end{small}
	\begin{minipage}{\linewidth}
		\footnotesize
		The high computational cost of BrainHarmonix-F (1,450 GFLOPs) and Brain-JEPA (1,272 GFLOPs) primarily derives from two factors: their support for long-range inputs exceeding 4k tokens and its large parameter count.
	\end{minipage}
	\vskip -0.05in
\end{table*}

Task-Specific Models:
\begin{itemize}
	\item EEGNet~\cite{lawhern2018eegnet} employs a compact CNN architecture with depthwise and separable convolutions to extract specialized features for EEG-based BCIs.
	\item EEGConformer~\cite{song2022eeg} is a compact convolutional Transformer architecture designed to extract both local and global representations for EEG decoding.
	\item ContraWR~\cite{yang2023self} is a CNN-based encoder that employs contrastive learning on short-time Fourier transform (STFT) spectrograms for automated EEG sleep staging.
	\item CNN-Transformer \cite{peh2022transformer} combines convolutional neural networks with transformer layers and employs a belief-matching loss to enable automated detection of EEG artifacts.
	\item FFCL \cite{li2022motor} employs two independent branches that fuse multilevel spatial–temporal features, integrating CNN and LSTM architectures for EEG decoding.
	\item ST-Transformer \cite{song2021transformer} is a transformer-based architecture that leverages spatial–temporal attention mechanisms to capture both spatial dependencies and temporal dynamics in EEG signals for decoding tasks.
	\item BrainNetTF \cite{kan2022brain} is a transformer-based graph model for fMRI brain network analysis, which employs an orthonormal clustering readout to capture structured network representations.
	\item BrainNetCNN \cite{kawahara2017brainnetcnn} employs specialized convolutional filters (edge-to-edge, edge-to-node, and node-to-graph) to capture the topological structure of brain networks for clinical outcome prediction.
\end{itemize}

Foundation Models:
\begin{itemize}
	\item LaBraM \cite{jiang2024large} is a unified EEG foundation model that enables cross-dataset generalization by segmenting raw EEG signals into channel-wise patches for transformer-based representation learning.
	\item CBraMod \cite{wang2024cbramod} is an EEG foundation model that employs a criss-cross attention mechanism to jointly capture frequency domain and temporal features for self-supervised representation learning.
	\item BrainHarmonix-F \cite{dong2025brain} is the fMRI branch of the BrainHarmonix, a multimodal brain foundation model designed to process and integrate both structural and functional MRI data.
	\item Brain-JEPA \cite{dong2024brain} is a fMRI foundation model that utilizes the Joint-Embedding Predictive Architecture (JEPA) alongside Brain Gradient Positioning for representation learning.
	\item BrainOmni \cite{xiao2025brainomni} is a brain foundation model pretrained on EEG and MEG recordings, incorporating learnable sensor position encoding and residual vector quantization.
\end{itemize}

To enable evaluation of these baselines on out-of-domain modalities, we adopt different adaptation strategies for task-specific models and pretrained foundation models.
Task-specific models (EEGNet, EEGConformer, etc.) carry no pretrained weights and are trained from scratch on every downstream task, in every modality. Their input layers can accordingly be reconfigured freely to the target number of channels or ROIs and the target sequence length, and every such model receives the same input. This includes the connectivity-based models: for BrainNetCNN and BrainNetTF, the connectivity matrix is computed inside the model's own input pipeline, so they consume the same raw inputs as all other baselines.
For pretrained brain foundation models (LaBraM, CBraMod, etc.), we instead aim to preserve the pretrained architecture and parameters as completely as possible. Rather than reconfiguring and reinitializing the input layer, we leave the architecture untouched and instead crop or pad the input signal to a shape the pretrained model accepts. For example, the inputs to LaBraM and CBraMod are adjusted to the length that is a multiple of their 200-sample patch size. These results should therefore be read as the transfer of a fully pretrained model to an out-of-domain input, with no pretrained parameters discarded. To verify that this procedure preserves useful information from pretraining rather than merely reusing the architecture, we conduct an additional control experiment on LEMON-EEG using BrainHarmonix-F, comparing the pretrained model adapted as described above against the identical architecture trained from scratch under the same downstream protocol. Removing the pretrained weights decreases BAC from \(77.95\) to \(73.13\), indicating that the pretrained representation provides transferable information even under this out-of-domain adaptation.

\section{Modality Importance Scores}
\label{sec:modality}
To quantify the contribution of each modality to overall model performance, we introduce two types of \emph{Modality Importance Scores}. The first, termed the Modality Shapley Value, is grounded in cooperative game theory and measures the marginal contribution of each modality. In this framework, each modality represents a "player," and the model's downstream performance represents the outcome of the "game". The second metric, termed the Relative Importance Score, is derived directly from modality ablation results in Section~\ref{modality_ablation}, which measures the relative performance degradation on downstream tasks when a specific modality is removed during pretraining.

Since downstream tasks employ evaluation metrics with different units and scales, we first normalize performance gains to ensure comparability. Formally, the normalized value function of a modality subset $S$ on downstream dataset $D_j$ is defined as:
\[
V_j(S) = \frac{P_j(S)-P_j(\emptyset)}{P_j(\mathcal{M})-P_j(\emptyset)}
\]
where $\mathcal{M} = \{M_{\text{fMRI}}, M_{\text{EEG}}, M_{\text{MEG}}\}$ is the set of all modalities used during joint pretraining, and $S \subseteq \mathcal{M}$ represents any subset (coalition) of these modalities. $P_j(S)$ denotes the downstream performance of the model pretrained on $S$, and $P_j(\emptyset)$ represents the baseline performance of a model trained from scratch (the empty coalition) on dataset $D_j$. By definition, $V_j(\emptyset) = 0$ indicates no pretraining gain, whereas $V_j(\mathcal{M}) = 1$ represents the total performance benefit derived from the omnifunctional joint pretraining. Higher values of $V_j(S)$ indicate greater gain derived from that specific subset of modalities. Notably, for lower-is-better metrics (e.g., MAE), the numerator and denominator are sign-flipped to preserve a consistent contribution direction.  

Based on this normalized value function, the Modality Shapley Value for modality $M_i$ on dataset $D_j$ is computed as its expected marginal contribution to the overall performance, weighted and summed over all possible modality coalitions:
\[\phi_{j}(M_i)
=
\sum_{S \subseteq \mathcal{M} \setminus \{M_i\}}
\frac{|S|!\left(|\mathcal{M}|-|S|-1\right)!}{|\mathcal{M}|!}
\left[
V_j(S \cup \{M_i\}) - V_j(S)
\right]
\]

The Shapley value inherently satisfies the cooperative game theory properties of symmetry and additivity, ensuring that the total pretraining gain is fully decomposed across modalities: 
\[
\phi_{j}(M_{\text{fMRI}}) + \phi_{j}(M_{\text{EEG}}) + \phi_{j}(M_{\text{MEG}}) = 1
\]

To complement this contribution analysis, the Relative Importance Score is introduced as a leave-one-out measure of modality dependence, quantifying the performance gain lost when modality $M_i$ is explicitly withheld.
\[
I_{M_i,D_j}
=
\frac{P_j(\mathcal{M})-P_j(\mathcal{M}\setminus\{M_i\})}
{P_j(\mathcal{M})-P_j(\emptyset)}
\times 100\%
\]
This metric evaluates how strongly the full Brain-OF depends on a single modality $M_i$ for task $D_j$. A positive score indicates that the modality contributes constructively to downstream performance. As above, lower-is-better metrics are adjusted to preserve consistent interpretation.

\section{Ablation Study of Mask Ratios in the MTFM}
To investigate the impact of the mask ratio in the MTFM, we evaluate ratios from 0.4 to 0.9 using the Brain-OF Base model on three representative downstream datasets: TUAB (EEG) for abnormal detection, ADNI (fMRI) for Alzheimer's diagnosis, and CamCAN (MEG) for brain age prediction. We adopt a symmetric masking strategy where the temporal and frequency mask ratios are set to the same value. The results are summarized in Table~\ref{maskratio}.
Performance drops sharply at a low mask ratio of 0.4 across all modalities and metrics, so we do not explore ratios below this threshold. For TUAB and ADNI, a ratio of 0.7 yields the strongest overall results (e.g., 81.88\% BAC on TUAB, 68.23\% BAC on ADNI), though 0.8 performs comparably on ADNI (74.53\% AUROC). In contrast, ratios of 0.6 and 0.5 are superior on CamCAN (e.g., 8.83 MAE at 0.6), but underperform on fMRI. Balancing these trade-offs, we select 0.7 as the final mask ratio for MTFM. Interestingly, different modalities exhibit preferences for distinct optimal ranges: fMRI favors higher ratios (0.7–0.8) whereas EEG and MEG benefit from moderate ones (0.5–0.7), revealing modality-specific sensitivities in self-supervised learning.

\begin{table*}[h]
	\vskip -0.1in
	\caption{Performance comparison across different mask ratios in the MTFM.}
	\label{maskratio}
	\centering
	\begin{scriptsize}
		\setlength{\tabcolsep}{1.6pt}
		\begin{tabular}{cccccccccc}
			\toprule
			\multirow{3}{*}{Mask Ratios} & \multicolumn{3}{c}{TUAB (EEG)} & \multicolumn{3}{c}{ADNI (fMRI)} & \multicolumn{3}{c}{CamCAN (MEG)} \\
			\cmidrule(r{0.02in}){2-4} \cmidrule(r{0.02in}){5-7} \cmidrule(r{0.02in}){8-10}
			& BAC(\%) $\uparrow$ & AUC-PR(\%) $\uparrow$ & AUROC(\%) $\uparrow$ & BAC(\%) $\uparrow$ & AUC-PR(\%) $\uparrow$ & AUROC(\%) $\uparrow$ & MAE $\downarrow$ & R2(\%) $\uparrow$ & MAPE(\%) $\downarrow$ \\
			\midrule
			$\mathcal{N}(0.4, 0.05^2)$ & 81.02{\fontsize{5pt}{0pt}\selectfont$\pm$0.12}& 87.17{\fontsize{5pt}{0pt}\selectfont$\pm$0.09}
			& 88.67{\fontsize{5pt}{0pt}\selectfont$\pm$0.06} & 61.10{\fontsize{5pt}{0pt}\selectfont$\pm$2.57}
			& 69.84{\fontsize{5pt}{0pt}\selectfont$\pm$2.36} & 67.54{\fontsize{5pt}{0pt}\selectfont$\pm$2.47}		& 11.93{\fontsize{5pt}{0pt}\selectfont$\pm$0.38} & 32.70{\fontsize{5pt}{0pt}\selectfont$\pm$3.92} 
			& 23.25{\fontsize{5pt}{0pt}\selectfont$\pm$0.77} \\
			$\mathcal{N}(0.5, 0.05^2)$ & 81.74{\fontsize{5pt}{0pt}\selectfont$\pm$0.21} & 87.59{\fontsize{5pt}{0pt}\selectfont$\pm$0.13} &	89.29{\fontsize{5pt}{0pt}\selectfont$\pm$0.19} & 65.01{\fontsize{5pt}{0pt}\selectfont$\pm$2.91} & 75.24{\fontsize{5pt}{0pt}\selectfont$\pm$2.28} &	71.42{\fontsize{5pt}{0pt}\selectfont$\pm$2.84} & 8.92{\fontsize{5pt}{0pt}\selectfont$\pm$0.18} & \textbf{58.19{\fontsize{5pt}{0pt}\selectfont$\pm$1.71}} &	17.00{\fontsize{5pt}{0pt}\selectfont$\pm$0.31}				\\
			$\mathcal{N}(0.6, 0.05^2)$ & 81.82{\fontsize{5pt}{0pt}\selectfont$\pm$0.31} & 87.64{\fontsize{5pt}{0pt}\selectfont$\pm$0.20} &	\textbf{89.45{\fontsize{5pt}{0pt}\selectfont$\pm$0.27}} & 63.74{\fontsize{5pt}{0pt}\selectfont$\pm$2.76} & 72.87{\fontsize{5pt}{0pt}\selectfont$\pm$3.12} &	70.85{\fontsize{5pt}{0pt}\selectfont$\pm$3.31} & \textbf{8.83{\fontsize{5pt}{0pt}\selectfont$\pm$0.19}} & 57.02{\fontsize{5pt}{0pt}\selectfont$\pm$1.94} &	\textbf{16.50{\fontsize{5pt}{0pt}\selectfont$\pm$0.48}}				 \\
			$\mathcal{N}(0.7, 0.05^2)$ & \textbf{81.88{\fontsize{5pt}{0pt}\selectfont$\pm$0.17}} &	\textbf{87.69{\fontsize{5pt}{0pt}\selectfont$\pm$0.11}} & 89.26{\fontsize{5pt}{0pt}\selectfont$\pm$0.09} & \textbf{68.23{\fontsize{5pt}{0pt}\selectfont$\pm$2.07}} & 76.23{\fontsize{5pt}{0pt}\selectfont$\pm$1.87} &	73.27{\fontsize{5pt}{0pt}\selectfont$\pm$0.71} & 8.99{\fontsize{5pt}{0pt}\selectfont$\pm$0.14} &	56.19{\fontsize{5pt}{0pt}\selectfont$\pm$1.84} &	16.92{\fontsize{5pt}{0pt}\selectfont$\pm$0.30}	  \\
			$\mathcal{N}(0.8, 0.05^2)$ & 81.51{\fontsize{5pt}{0pt}\selectfont$\pm$0.09} &	87.46{\fontsize{5pt}{0pt}\selectfont$\pm$0.06} & 89.14{\fontsize{5pt}{0pt}\selectfont$\pm$0.09} & 67.86{\fontsize{5pt}{0pt}\selectfont$\pm$2.62} &	\textbf{76.77{\fontsize{5pt}{0pt}\selectfont$\pm$3.17}} &	\textbf{74.53{\fontsize{5pt}{0pt}\selectfont$\pm$4.71}} & 9.45{\fontsize{5pt}{0pt}\selectfont$\pm$0.40} &	53.27{\fontsize{5pt}{0pt}\selectfont$\pm$3.62} & 18.06{\fontsize{5pt}{0pt}\selectfont$\pm$0.51}	\\
			$\mathcal{N}(0.9, 0.05^2)$ & 81.26{\fontsize{5pt}{0pt}\selectfont$\pm$0.25} & 87.31{\fontsize{5pt}{0pt}\selectfont$\pm$0.16} & 88.76{\fontsize{5pt}{0pt}\selectfont$\pm$0.35} & 64.76{\fontsize{5pt}{0pt}\selectfont$\pm$3.78} & 71.61{\fontsize{5pt}{0pt}\selectfont$\pm$3.31} & 69.25{\fontsize{5pt}{0pt}\selectfont$\pm$2.23} & 10.39{\fontsize{5pt}{0pt}\selectfont$\pm$0.32} & 44.38{\fontsize{5pt}{0pt}\selectfont$\pm$2.75} & 19.94{\fontsize{5pt}{0pt}\selectfont$\pm$0.66} \\
			\bottomrule
		\end{tabular}
	\end{scriptsize}
	\begin{minipage}{\linewidth}
		\footnotesize
		The results presented above are obtained with the Brain-OF Base.
	\end{minipage}
	\vskip -0.2in
\end{table*}

\section{Ablation Study of Router Bias Update Rate $\gamma$}
To examine the sensitivity of Brain-OF to the router bias update rate $\gamma$ during pretraining, we conduct an ablation study over a range of values from $10^{-5}$ to $10^{-2}$. Each pretrained model is subsequently fully finetuned and evaluated on three representative downstream tasks: TUAB (EEG), ADNI (fMRI) and CamCAN (MEG).

\begin{figure}[h]
	\centering
		\includegraphics[width=\columnwidth]{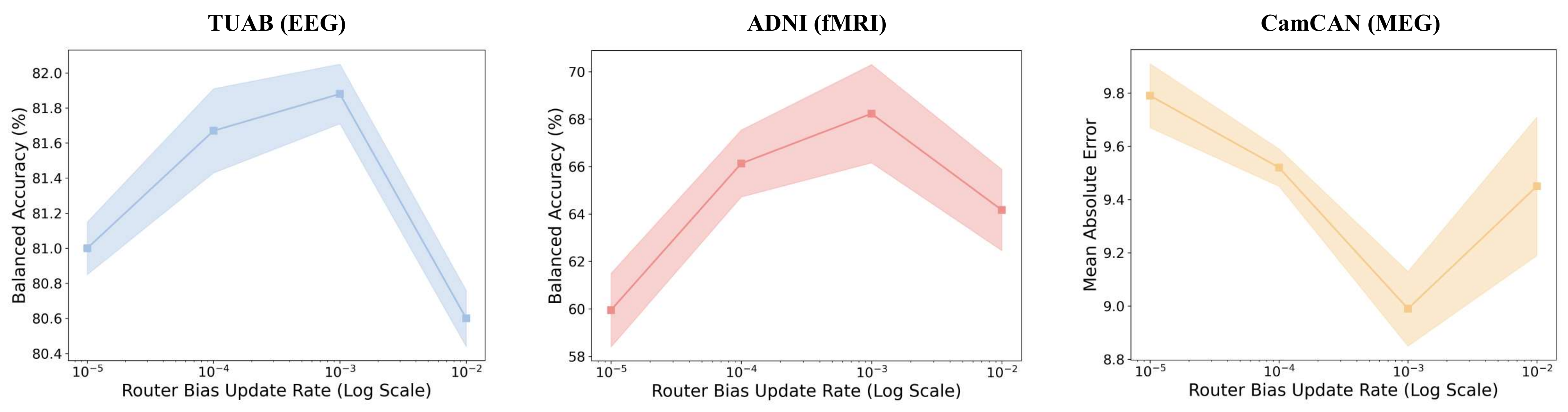}
		\caption{
			The impact of the router bias update rate $\gamma$ during pretraining. The performances are evaluated on downstream tasks across three modalities.
		}
		\label{bias_update_rate}
\end{figure}

As illustrated in Figure~\ref{bias_update_rate}, a consistent optimal configuration emerges across all modalities. In particular, $\gamma = 10^{-3}$ achieves the highest Balanced Accuracy on EEG and fMRI tasks and the lowest Mean Absolute Error on MEG. Larger values (e.g., $10^{-2}$) lead to noticeable performance degradation, likely due to overly aggressive bias updates that destabilize expert routing and impair specialization. In contrast, smaller values ($10^{-5}$) fail to sufficiently update the router bias, limiting the effectiveness of load balancing.

Interestingly, we observe a different preference during the finetuning stage. While $\gamma = 10^{-3}$ is crucial for effective pretraining, substantially smaller update rates (e.g., $\gamma = 10^{-5}$) generally yield improved stability and performance during downstream finetuning, preserving the specialized knowledge acquired during pretraining.

\section{Task-State fMRI Downstream Evaluation}
To further evaluate Brain-OF on fMRI modality, we extend downstream evaluation to two task-state fMRI benchmarks: HAD and BOLD5000. Both datasets involve visual stimuli tasks. Specifically, HAD is a complex 180-way classification of human actions based on evoked fMRI responses (evaluated via Top-10 Accuracy). BOLD5000 is a 3-way cognitive decoding task to identify the source domain of visual stimuli (images from COCO, ImageNet, or Scene databases, evaluated via Balanced Accuracy).

As shown in Table~\ref{tab:taskfmri}, Brain-OF achieves stronger performance than existing fMRI foundation models on both datasets. Notably, on the highly complex HAD benchmark, Brain-OF Base achieves 16.17\% Top-10 Accuracy, surpassing Brain-JEPA by 3.86\%. Similarly, it also achieves a better performance on BOLD5000 (40.49\% BAC). This empirical evidence demonstrates that Brain-OF does not merely treat fMRI as a static clinical biomarker, but possesses robust, event-related functional decoding capabilities.

\begin{table*}[htbp]
	\caption{Additional downstream evaluation on task-state fMRI.}
	\label{tab:taskfmri}
	\centering
	\begin{tabular}{lccc}
		\toprule
		\textbf{Methods} & \textbf{BrainHarmonix-F} & \textbf{Brain-JEPA} & \textbf{Brain-OF Base} \\
		
		\midrule
		\textbf{HAD} (Top-10 Acc \%) $\uparrow$  &	6.38$\pm$0.16 &	12.31$\pm$1.69 & \textbf{16.17$\pm$0.44} \\
		\textbf{BOLD5000} (BAC \%) $\uparrow$ & 38.94$\pm$1.66 & 39.79$\pm$0.41 & \textbf{40.49$\pm$0.67}	\\
		\bottomrule
	\end{tabular}
	\vskip -0.1in
\end{table*}

\section{Visualization of Reconstruction}
We visualize the reconstruction quality of Brain-OF Huge to assess the effectiveness of the proposed Masked Temporal-Frequency Modeling (MTFM) objective across fMRI, EEG, and MEG. Beyond simple waveform reconstruction, we examine whether the model preserves critical high-order features.

\begin{figure}[h]
	\vskip -0.15in
	\centering
		\includegraphics[width=\columnwidth]{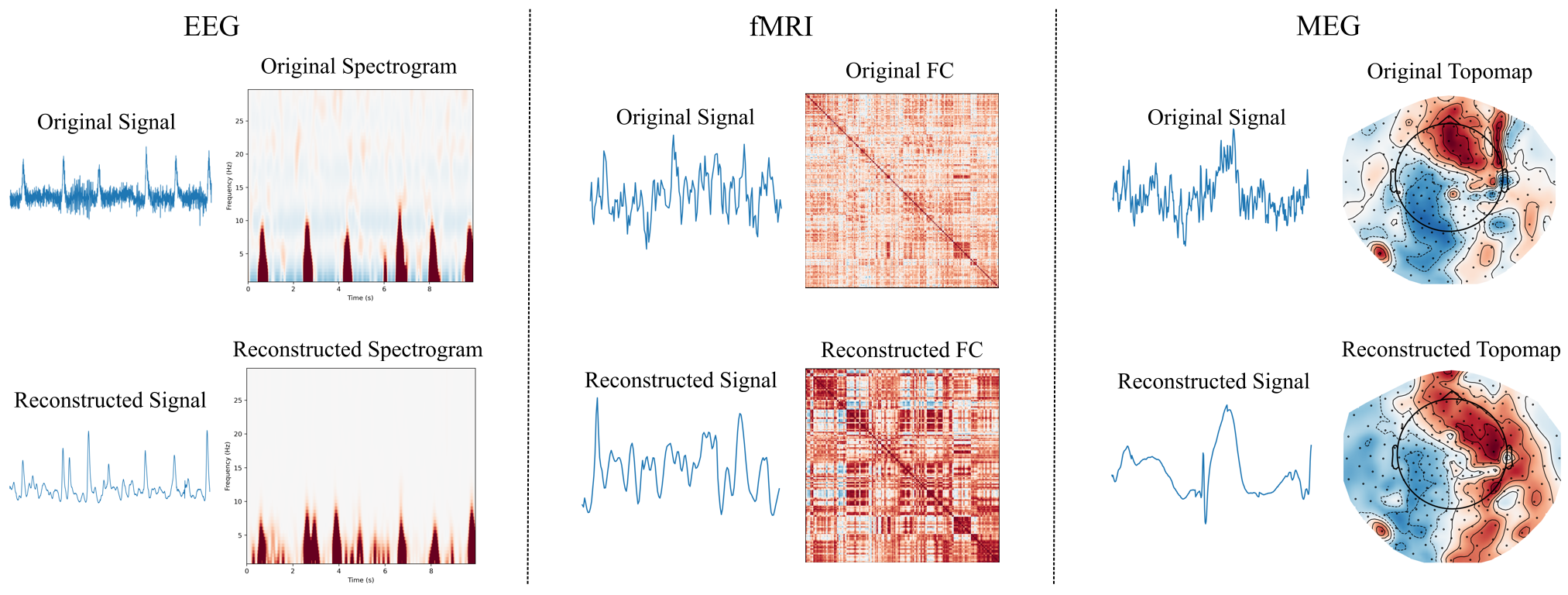}
		
		\caption{Qualitative visualization of signal reconstruction across fMRI, EEG and MEG.}
		\label{visualization_rec}
\end{figure}

As shown in Figure~\ref{visualization_rec}, Brain-OF achieves high-fidelity reconstruction with effective noise suppression. For EEG, reconstructed signals closely match the original temporal dynamics while preserving dominant spectral bursts. In fMRI, functional connectivity derived from the original and reconstructed BOLD signals show that the complex inter-regional correlation structure is well preserved. For MEG, reconstructed topomaps accurately recover localized scalp-level magnetic field patterns despite signal sparsity.
Overall, these results indicate that Brain-OF learns meaningful spatiotemporal and spectral generative structure rather than memorizing raw inputs.

\section{Additional Comparison with BrainOmni}
We provide an additional comparison between Brain-OF Base and the recent foundation model BrainOmni Base, which is pretrained on EEG and MEG signals. BrainOmni is not included in the main comparison because its heavy reliance on rigid EMEG sensor layouts makes it structurally difficult to evaluate on fMRI datasets. In addition, its pretraining corpus includes the CamCAN dataset, which gives it an unfair "seen-data" advantage on downstream tasks. Nevertheless, we include it here for completeness. 

As shown in Table~\ref{tab:brainomni}, Brain-OF achieves a +4.52\% improvement on CHB-MIT and better performance on CamCAN-MEG (MAE 8.99 vs. 11.99). On TUAB, both models demonstrate highly comparable performance (81.88\% vs. 81.90\%). These results further support the robustness and generalization of Brain-OF across heterogeneous neuroimaging modalities.

\begin{table*}[htbp]
	\vskip -0.1in
	\caption{Additional comparison between Brain-OF and BrainOmni.}
	\label{tab:brainomni}
	\centering
	\begin{tabular}{lccc}
		\toprule
		\multirow{4}{*}{\textbf{Methods}} & \textbf{CHB-MIT} & \textbf{TUAB} & \textbf{CamCAN} \\
		\cmidrule(r{0.02in}){2-4}
		& \textbf{EEG} & \textbf{EEG} & \textbf{MEG} \\
		\cmidrule(r{0.02in}){2-4}
		& BAC(\%) $\uparrow$ & BAC(\%) $\uparrow$ & MAE $\downarrow$ \\
		\midrule
		BrainOmni Base  & 69.47$\pm$2.19 & \textbf{81.90$\pm$0.50} & 11.99$\pm$0.84  \\
		Brain-OF Base & \textbf{73.99$\pm$1.48} & 81.88$\pm$0.17 & \textbf{8.99$\pm$0.14}	\\
		\bottomrule
	\end{tabular}
	\vskip -0.1in
\end{table*}

\section{Multimodal Fusion Comparison with Adapted BrainHarmonix}
A direct comparison with existing multimodal brain foundation models is challenging because, to the best of our knowledge, no publicly available model provides a directly comparable pretrained pipeline for fusing fMRI with EEG or MEG on paired downstream datasets. Existing multimodal foundation models typically focus on different modality combinations. BrainHarmonix jointly models structural MRI (sMRI) and fMRI, BrainOmni jointly pretrains on EEG and MEG, while BIOT models heterogeneous biosignals such as EEG and ECG. These settings therefore do not directly address fusion between hemodynamic fMRI and electrophysiological EEG/MEG signals.

To provide a multimodal comparison, we adapt the BrainHarmonix fusion architecture, which originally consists of dual sMRI--fMRI branches, to our target setting. Specifically, we replace the original sMRI branch with a duplicated fMRI branch, such that one stream receives fMRI while the other receives either MEG or EEG. The complete two-branch model and fusion module are then finetuned end-to-end on the corresponding paired downstream training set. Because this configuration differs from the original sMRI--fMRI setting of BrainHarmonix, we regard it as an adapted architecture-level reference rather than the official BrainHarmonix configuration.

As shown in Table~\ref{tab:multimodal_baseline}, the adapted BrainHarmonix fusion does not outperform its strongest unimodal branch on either dataset: its multimodal performance is worse than MEG alone on CamCAN and worse than EEG alone on LEMON. In contrast, Brain-OF Base and Large improve over their corresponding strongest unimodal settings on both benchmarks. Brain-OF also consistently outperforms the adapted BrainHarmonix baseline in the multimodal setting. These results suggest that a fusion mechanism originally designed for sMRI--fMRI integration does not directly transfer to heterogeneous fMRI--electrophysiology fusion, highlighting the distinct challenge addressed by Brain-OF.

\begin{table*}[h]
	\vskip -0.1in
	\caption{Comparison of multimodal fusion performance on CamCAN and LEMON. BrainHarmonix denotes an adapted architecture in which the original sMRI branch is replaced with a duplicated fMRI branch.}
	\label{tab:multimodal_baseline}
	\centering
	\setlength{\tabcolsep}{4.3pt}
	\begin{small}
	
	\begin{tabular}{lcccccc}
		\toprule
		\multirow{4}{*}{\textbf{Methods}} & \multicolumn{3}{c}{\textbf{CamCAN}} & \multicolumn{3}{c}{\textbf{LEMON}} \\
		\cmidrule(r{0.02in}){2-4} \cmidrule(r{0.02in}){5-7}
		& \textbf{fMRI} & \textbf{MEG} & \textbf{Multimodal}& \textbf{fMRI} & \textbf{EEG} & \textbf{Multimodal} \\
		\cmidrule(r{0.02in}){2-4} \cmidrule(r{0.02in}){5-7}
		& MAE $\downarrow$ & MAE $\downarrow$ & MAE $\downarrow$ & BAC(\%) $\uparrow$ & BAC(\%) $\uparrow$ & BAC(\%) $\uparrow$ \\
		\midrule
		BrainHarmonix
		& 14.81 $\pm$ 0.49
		& 8.73 $\pm$ 0.09
		& 11.61 $\pm$ 0.75
		& 63.11 $\pm$ 1.32
		& 77.95 $\pm$ 2.36
		& 74.43 $\pm$ 0.98 \\
		
		Brain-OF Base
		& 12.09 $\pm$ 0.78
		& 8.99 $\pm$ 0.10
		& \textbf{8.87 $\pm$ 0.09}
		& 58.95 $\pm$ 2.13
		& 76.05 $\pm$ 0.44
		& \textbf{76.77 $\pm$ 0.60} \\
		
		Brain-OF Large
		& 11.71 $\pm$ 0.67
		& 8.86 $\pm$ 0.21
		& \textbf{8.78 $\pm$ 0.07}
		& 62.85 $\pm$ 4.29
		& 78.04 $\pm$ 1.59
		& \textbf{78.32 $\pm$ 1.64} \\
		
		Brain-OF Huge
		& \textbf{9.51 $\pm$ 0.51}
		& \textbf{7.87 $\pm$ 0.22}
		& 8.21 $\pm$ 0.11
		& 61.79 $\pm$ 1.29
		& \textbf{80.39 $\pm$ 0.59}
		& \textbf{81.10 $\pm$ 0.54} \\
		
		\bottomrule
	\end{tabular}
	\end{small}
	\vskip -0.1in
\end{table*}

\section{Hyperparameter Configurations}
\label{hyperparameters}
In this section, we provide the detailed hyperparameter configurations in the Brain-OF pretraining, as well as the optimization settings for downstream tasks. All models were implemented in PyTorch and trained on NVIDIA A100 GPUs using automatic mixed precision.

\subsection{Pretraining Settings}
\begin{table}[H]
	\vskip -0.1in
	\caption{Hyperparameters for pretraining Brain-OF.}
	\label{pretraining_hyperparameter}
	\vskip -0.05in
	\centering
		\begin{small}
		\setlength{\tabcolsep}{5.2pt}
			\begin{tabular}{ccccc}
				\toprule
				\multicolumn{2}{c}{\textbf{Hyperparameters}}
				& \textbf{Brain-OF Base} & \textbf{Brain-OF Large} & \textbf{Brain-OF Huge} \\
				\midrule
				\multicolumn{2}{c}{Model Size} & 47.5M & 331M & 1.7B \\
				\multicolumn{2}{c}{Activated Parameters} & 21.5M & 150M & 500M \\
				\multicolumn{2}{c}{Patch size} & \multicolumn{3}{c}{64} \\
				\multicolumn{2}{c}{Max sequence length} & \multicolumn{3}{c}{2048} \\
				\multicolumn{2}{c}{Semantic space size} & \multicolumn{3}{c}{128} \\
				\multicolumn{2}{c}{Padding token} & \multicolumn{3}{c}{Full zero} \\
				\midrule
				\multirow{5}{*}{Brain Signal Encoder} & Input dimension & \{1, 4, 4\} & \{1, 8, 8\} & \{1, 12, 12\} \\
				& Output dimension & \{4, 4, 4\} & \{8, 8, 8\} & \{12, 12, 12\} \\
				& Kernel size & \multicolumn{3}{c}{\{3, 3, 3\}} \\
				& Stride & \multicolumn{3}{c}{\{1, 1, 1\}}\\
				& Padding & \multicolumn{3}{c}{\{1, 1, 1\}} \\
				\midrule
				\multirow{7}{*}{Brain-OF} 
				& ARNESS layer & 4 & 4 & 6 \\
				& Backbone layer & 12 & 24 & 36 \\
				& Attention head & 4 & 8 & 12\\
				& Hidden dimension & 256 & 512 & 768 \\
				& Routed expert & 8 & 16 & 28 \\
				& Activated expert & 2 & 4 & 4 \\
				& Shared expert & \multicolumn{3}{c}{1}\\
				\midrule
				\multirow{13}{*}{Pretraining settings} 
				& Epochs & \multicolumn{3}{c}{150} \\
				& \multirow{2}{*}{Global batch size} & 7,168  & 7,040  & 7,168  \\ && (224$\times$32 GPUs) & (80$\times$88 GPUs) & (32$\times$224 GPUs)\\
				& Optimizer & \multicolumn{3}{c}{AdamW} \\
				& Adam $\beta$ & \multicolumn{3}{c}{(0.9, 0.95)} \\
				& Peak learning rate & \multicolumn{3}{c}{3e-3} \\
				& Warmup learning rate & \multicolumn{3}{c}{3e-5} \\
				& Router bias update rate & \multicolumn{3}{c}{1e-3} \\
				& Learning rate scheduler & \multicolumn{3}{c}{Cosine}\\
				& Warmup epochs & \multicolumn{3}{c}{8} \\
				& Weight decay & \multicolumn{3}{c}{0.05}\\
				& Gradient clipping & \multicolumn{3}{c}{5.0} \\
				& Mask ratio & \multicolumn{3}{c}{$\mathcal{N}(0.7, 0.05^2)$} \\
				& Layer scale init & 1e-4 & 1e-5 & 1e-6\\
				\bottomrule
			\end{tabular}
		\end{small}
	\vskip -0.1in
\end{table}

\subsection{Finetuning Settings on Downstream Tasks}
For downstream tasks, Brain-OF is fully finetuned with a lightweight task-specific linear head following global average pooling. We employ cross-entropy loss for classification tasks and mean squared error loss for regression tasks. Optimization is performed using AdamW with $\beta=(0.9, 0.99)$ and a cosine learning rate scheduler. Detailed finetuning hyperparameter configurations for different model variants are provided in Table~\ref{finetune_params}, Table~\ref{finetune_params_large} and Table~\ref{finetune_params_huge}.
\begin{table}[H]
	\vskip -0.1in
	\centering
	\caption{Hyperparameters for downstream finetuning (Brain-OF Base).}\vskip -0.05in
	\label{finetune_params}
		\begin{small}
			\setlength{\tabcolsep}{2.8pt}
			\begin{tabular}{lcccccccc}
				\toprule
				\textbf{Datasets} & \textbf{Global bsz} & \textbf{Epochs} & \textbf{Warmup epochs} &\textbf{Lr} & \textbf{Warmup lr}& \textbf{Wd} & \textbf{Drop out} & \textbf{Drop path}\\
				\midrule
				SEED-V & 1,024 & 50 & 5 & 1e-4 & 1e-6 & 5.0 & 0.0 & 0.0\\
				CHB-MIT & 1,024 & 50 & 5 & 3e-5 & 3e-7 & 0.1 & 0.0 & 0.1 \\
				TUAB & 1,024 & 50 & 5 & 7e-6 & 7e-8 & 0.1 & 0.0 & 0.0 \\
				ADNI & 32 & 50 & 5 & 1.5e-4 & 1.5e-6 & 0.1 & 0.1 & 0.0\\
				ADHD-200 & 32 & 50 & 5 & 3e-5 & 3e-7 & 0.01 & 0.1 & 0.0 \\
				CamCAN-fMRI & 32 & 100 & 5 & 7.5e-4 & 7.5e-6 & 0.1 & 0.1 & 0.0 \\
				CamCAN-MEG & 1,024 & 50 & 5 & 1e-4 & 1e-6 & 5e-4 & 0.0 & 0.0 \\
				CamCAN-MM & 512 & 50 & 5 & 2e-4 & 2e-6 &0.1 & 0.0 & 0.2 \\
				LEMON-fMRI & 32 & 100 & 5 & 2e-4 & 2e-6 & 0.1 & 0.1 & 0.1 \\
				LEMON-EEG & 512 & 50 & 5 & 1.5e-5 & 1.5e-7 & 0.1 & 0.0 & 0.0\\
				LEMON-MM & 512 & 50 & 5 & 4e-5 & 4e-7 & 0.1 & 0.0 & 0.3 \\
				\bottomrule
			\end{tabular}
			\vskip 0.01in
			\begin{minipage}{\linewidth}
				\footnotesize
				MM (Multimodal): Indicates finetuning Brain-OF on multimodal scenarios. Wd: Stands for weight decay. Specifically, CamCAN comprises paired fMRI and MEG, while LEMON comprises paired fMRI and EEG.
			\end{minipage}
		\end{small}
\end{table}

\begin{table}[H]
	\vskip -0.1in
	\centering
	\caption{Hyperparameters for downstream finetuning (Brain-OF Large).}\vskip -0.05in
	\label{finetune_params_large}
		\begin{small}
			\setlength{\tabcolsep}{3pt}
			\begin{tabular}{lcccccccc}
				\toprule
				\textbf{Datasets} & \textbf{Global bsz} & \textbf{Epochs} & \textbf{Warmup epochs} &\textbf{Lr} & \textbf{Warmup lr}& \textbf{Wd} & \textbf{Drop out} & \textbf{Drop path}\\
				\midrule
				SEED-V & 1,024 & 50 & 5 & 4e-4 & 4e-6 & 5.0 & 0.0 & 0.1 \\
				CHB-MIT & 1,024 & 50 & 5 & 3e-5 & 3e-7 & 0.5 & 0.0 & 0.1 \\
				TUAB & 1,024 & 50 & 5 & 3e-5 & 3e-7 & 0.1 & 0.0 & 0.3 \\
				ADNI & 32 & 50 & 5 & 1e-4 & 1e-6 & 0.1 & 0.0 & 0.3 \\
				ADHD-200 & 32 & 50 & 5 & 4e-5 & 4e-7 & 0.01 & 0.1 & 0.2 \\
				CamCAN-fMRI & 32 & 100 & 5 & 3e-3 & 3e-5 & 0.1 & 0.1 & 0.1 \\
				CamCAN-MEG & 1,024 & 50 & 5 & 6e-4 & 6e-6 & 0.1 & 0.0 & 0.3 \\
				CamCAN-MM & 512 & 50 & 5 & 8e-5 & 8e-7 & 0.05 & 0.0 & 0.2 \\
				LEMON-fMRI & 32 & 100 & 5 & 3e-4 & 3e-6 & 0.1 & 0.0 & 0.1 \\
				LEMON-EEG & 512 &  50 & 5 & 6e-4 & 6e-6 & 0.1 & 0.0 & 0.1 \\
				LEMON-MM & 512 & 50 & 5 & 4e-4 & 4e-6 & 0.1 & 0.0 & 0.1 \\
				\bottomrule
			\end{tabular}
			\vskip 0.01in
			\begin{minipage}{\linewidth}
				\footnotesize
				MM (Multimodal): Indicates finetuning Brain-OF on multimodal scenarios. Wd: Stands for weight decay. Specifically, CamCAN comprises paired fMRI and MEG, while LEMON comprises paired fMRI and EEG.
			\end{minipage}
		\end{small}
	\vskip -0.15in
\end{table}

\begin{table}[H]
	\vskip -0.1in
	\centering
	\caption{Hyperparameters for downstream finetuning (Brain-OF Huge).}
	\vskip -0.05in
	\label{finetune_params_huge}
		\begin{small}
			\setlength{\tabcolsep}{3pt}
			\begin{tabular}{lcccccccc}
				\toprule
				\textbf{Datasets} & \textbf{Global bsz} & \textbf{Epochs} & \textbf{Warmup epochs} &\textbf{Lr} & \textbf{Warmup lr}& \textbf{Wd} & \textbf{Drop out} & \textbf{Drop path}\\
				\midrule
				SEED-V & 1,024 & 50 & 5 & 1e-4 & 1e-6 & 5.0 & 0.0 & 0.1 \\
				CHB-MIT & 1,024 & 50 & 5 & 1e-4 & 1e-6 & 0.1 & 0.0 & 0.1 \\
				TUAB & 1,024 & 50 & 5 & 7e-6 & 7e-8 & 0.1 & 0.0 & 0.1 \\
				ADNI & 32 & 50 & 5 & 3e-4 & 3e-6 & 0.1 & 0.0 & 0.1 \\
				ADHD-200 & 32 & 50 & 5 & 1e-4 & 1e-6 & 0.1 & 0.0 & 0.3 \\
				CamCAN-fMRI & 32 & 100 & 5 & 5e-3 & 5e-5 & 0.1 & 0.0 & 0.1 \\
				CamCAN-MEG & 1,024 & 50 & 5 & 4e-4 & 4e-6 & 0.1 &0.0 & 0.3 \\
				CamCAN-MM & 512 & 50 & 5 & 1e-3 & 1e-4 & 0.1 & 0.0 & 0.3\\
				LEMON-fMRI & 32 & 100 & 5 & 5e-4 & 5e-6 & 0.1 & 0.001 & 0.1 \\
				LEMON-EEG & 512 & 50& 5 & 1.5e-4 & 1.5e-6 & 0.1 & 0.0 & 0.1\\
				LEMON-MM & 512 & 50 & 5 & 4e-4 & 4e-6 & 0.1 &0.0 & 0.1\\
				\bottomrule
			\end{tabular}
			\vskip 0.01in
			\begin{minipage}{\linewidth}
				\footnotesize
				MM (Multimodal): Indicates finetuning Brain-OF on multimodal scenarios. Wd: Stands for weight decay. Specifically, CamCAN comprises paired fMRI and MEG, while LEMON comprises paired fMRI and EEG.
			\end{minipage}
		\end{small}
	\vskip -0.15in
\end{table}

\section{Efficiency Analysis}
\label{sec: efficiency}
\vskip -0.05in
We evaluate the computational efficiency of Brain-OF across three model scales: Base, Large and Huge. Table~\ref{efficiency_analysis} summarizes the parameter counts, training resources and inference latency.

\textbf{Model Scaling and Training Cost.}
Brain-OF employs a Sparse Mixture of Experts architecture, which decouples the total parameter count from the computational cost per forward pass. Although Brain-OF Huge contains 1.7B parameters, only approximately 500M parameters (29\%) are activated per token. This sparsity enables efficient scaling: while the total parameter count increases by $35\times$ from Base (47.5M) to Huge (1.7B), the number of active parameters grows by only $\sim23\times$. Pretraining the Huge variant required 153 hours on 224 NVIDIA A100 GPUs.

\textbf{Inference Latency and Throughput.}
Inference latency is measured on a single NVIDIA A100 GPU using BF16 precision, with a sequence length of 2,048 tokens and a batch size of 16. The average \emph{per-sample} latency remains low: 5.5 ms for Base, 11.1 ms for Large, and 18.6 ms for Huge. This corresponds to a throughput of 181.8 samples/s for the Base model, indicating that Brain-OF remains suitable for real-time despite its large-scale architecture. Notably, the efficiency is further enhanced by ARNESS, which compresses high-dimensional brain signals into a compact semantic representation, significantly reducing the effective token length and computational burden.
\begin{table*}[h]
	\vskip -0.05in
	\caption{Efficiency analysis across model variants. \textbf{Active Params} denotes the number of parameters used per forward pass (due to sparse MoE). \textbf{Inference Latency} is reported for a batch size of 16 with sequence length 2,048 on a single A100 GPU.}
	\label{efficiency_analysis}
	\centering
		\begin{small}
			\setlength{\tabcolsep}{5.5pt}
			\begin{tabular}{lccccccc}
				\toprule
				\multirow{2}{*}{\textbf{Model Variants}} & \textbf{Total} & \textbf{Active} & \textbf{Comp.} & \multicolumn{2}{c}{\textbf{Pretraining Cost}} & \multicolumn{2}{c}{\textbf{Inference (Batch=16)}} \\
				\cmidrule(lr){5-6} \cmidrule(lr){7-8}
				& \textbf{Params} & \textbf{Params} & \textbf{GFLOPs} & \textbf{Time} & \textbf{\# GPUs} & \textbf{Latency} & \textbf{Throughput} \\
				\midrule
				Brain-OF Base & 47.5M & 21.5M & 11.29 & 71h 50m & 32 & 88 ms & 181.8 samples/s\\
				Brain-OF Large & 331M & 150M & 61.01 & 82h 06m & 88 & 178 ms & 89.9 samples/s\\
				Brain-OF Huge & 1.7B & 500M & 178.19 & 153h 03m & 224 & 297 ms & 53.8 samples/s\\
				\bottomrule
			\end{tabular}
		\end{small}
	\vskip -0.15in
\end{table*}

\section{Representation Visualizations on Downstream Datasets}
\label{sec:vis_representation}
\vskip -0.1in

\begin{figure}[h]
	\vskip -0.15in
	\centering
		\includegraphics[width=\columnwidth]{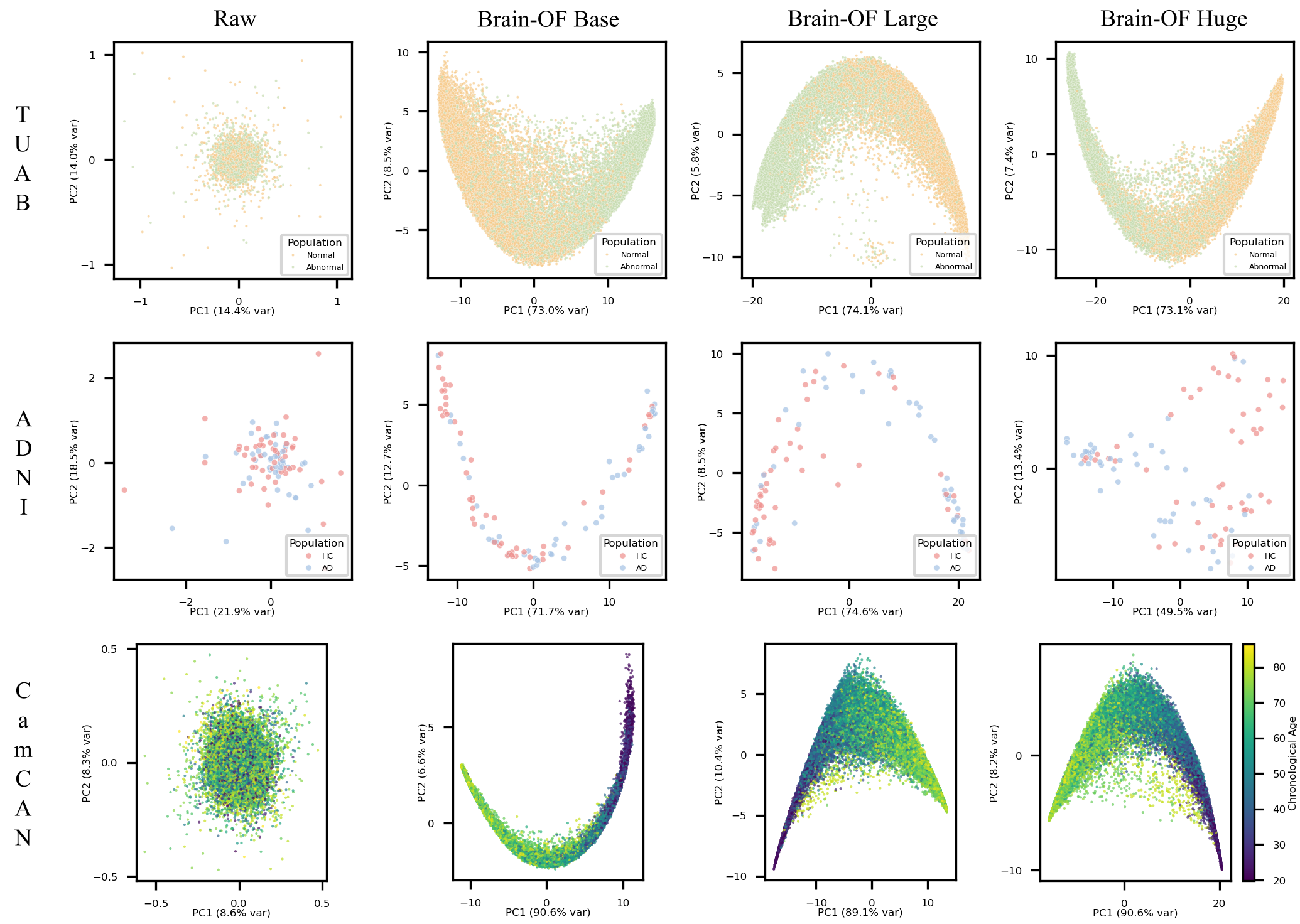}
		\vskip -0.1in
		\caption{\textbf{PCA visualization of learned representations across model scales.} Comparison of raw input features (Left) with representations from Brain-OF Base, Large and Huge (Right). Rows correspond to different modalities and tasks: \textbf{Top:} TUAB (EEG, binary classification); \textbf{Middle:} ADNI (fMRI, binary classification); \textbf{Bottom:} CamCAN (MEG, age regression). As model size increases, the representations exhibit clearer class separation (Top/Middle) and a more coherent continuous age-related trajectory for regression (Bottom), demonstrating the benefits of scaling.}
		\label{vis_representation}
		\vskip -0.3in
\end{figure}

To qualitatively assess the learned representations, we employ principal component analysis (PCA)~\cite{pearson1901liii} to visualize the feature embeddings generated by finetuned Brain-OF across three downstream datasets. Figure~\ref{vis_representation} illustrates the evolution of representations from raw input features (Column~1) to embeddings learned by Brain-OF Base, Large and Huge (Columns~2–4).

The first two rows correspond to classification tasks on TUAB (EEG) and ADNI (fMRI). In the raw feature space, samples from different classes (e.g., Normal vs.\ Abnormal, Healthy Control vs.\ Alzheimer’s Disease) are heavily entangled. As model capacity increases from Base to Huge, a clear trend emerges: class-specific clusters become progressively more compact and better separated.
On TUAB, the first principal component of Brain-OF Base explains \(73.0\%\) of the embedding variance and is strongly associated with the binary class label (Pearson \(r=0.701\); Spearman \(\rho=0.678\)), achieving an AUC of \(0.892\) using this component alone. This correlation becomes slightly stronger for Brain-OF Large (\(r=0.707\), \(\rho=0.687\), AUC \(=0.898\)) and Brain-OF Huge (\(r=0.723\), \(\rho=0.700\), AUC \(=0.905\)). On ADNI, the class-related structure is weaker but remains evident. The corresponding statistics improve from Brain-OF Base (\(r=0.380\), \(\rho=0.388\), AUC \(=0.724\)) to Large (\(r=0.478\), \(\rho=0.438\), AUC \(=0.753\)) and Huge (\(r=0.484\), \(\rho=0.496\), AUC \(=0.786\)). These results indicate that scaling the model capacity directly enhances its ability to extract discriminative, class-specific representations.
The third row visualizes the regression task on CamCAN (MEG), where color gradients indicate chronological age. For Brain-OF Base, the first principal component exhibits a strong association with age (\(r=0.746\); \(\rho=0.735\)). The age association further increases for Brain-OF Large (\(r=0.784\), \(\rho=0.762\)) and Brain-OF Huge (\(r=0.805\), \(\rho=0.772\)). Rather than forming discrete clusters, the CamCAN embeddings exhibit a continuous trajectory consistent with the continuous nature of chronological age. This structure suggests that Brain-OF captures meaningful age-related biological variation in the MEG signals.
Overall, the results further show that scaling Brain-OF generally improves class separability in classification tasks and strengthens the organization of continuous age-related variation in regression, supporting the quality and scalability of the learned representations. 

\section{Linear Probing and Pretraining Controls}
\label{sec:linear_probing}

Beyond the representation space visualization in Appendix~\ref{sec:vis_representation}, we further evaluate the quality of the learned representations through frozen-backbone linear probing and same-architecture no-pretraining controls. We conduct these analyses on three representative downstream tasks spanning EEG (TUAB), fMRI (ADNI), and MEG (CamCAN).

We freeze the entire pretrained backbone and train only a prediction head. As shown in Table~\ref{tab:linear_probing}, on both classification tasks, linear probing outperforms full finetuning of the same architecture from random initialization. Specifically, on TUAB, linear probing achieves 79.14\% vs. 77.59\% (Base) and 81.92\% vs. 80.82\% (Huge). On ADNI, the corresponding results are 65.22\% vs. 63.39\% and 70.76\% vs. 64.85\%. Since the from-scratch models have identical architectures and strictly more trainable parameters, these results demonstrate that the pretrained representations themselves contain substantial task-relevant information beyond the effect of model capacity. 

The no-pretraining control uses an identical architecture trained from scratch, thereby holding model capacity fixed while varying only the availability of pretrained weights. Pretrained Brain-OF Base outperforms the larger Brain-OF Huge trained from scratch on both TUAB (81.88 vs.\ 80.82) and ADNI (68.23 vs.\ 64.85). At matched model scale, removing pretraining decreases BAC by 4.29\% and 4.84\% on TUAB and ADNI for Brain-OF Base, and by 2.05\% and 6.94\% for Brain-OF Huge. The pretraining gains on TUAB and ADNI are also statistically significant at both model scales (two-sided Welch's $t$-test, corrected $p<0.01$). These results indicate that simply increasing model capacity does not recover the gains provided by pretraining. On CamCAN-MEG, however, the effect is less straightforward. Full finetuning from pretrained checkpoints still yields numerically lower MAE than training from scratch (8.99 vs.\ 9.69 for Base; 7.87 vs.\ 8.59 for Huge), but these differences are not statistically significant, and frozen linear probing performs worse than full training from scratch. This suggests that the pretrained MEG representations remain useful for downstream adaptation, but brain-age regression benefits more strongly from nonlinear backbone adaptation than from a linear readout alone.

\begin{table}[h]
	\centering
	\caption{Comparison of full finetuning, frozen-backbone linear probing, and training from scratch for Brain-OF Base and Huge.}
	\label{tab:linear_probing}
	\begin{small}
	\begin{tabular}{lcccc}
		\toprule
		\multirow{3}{*}{\textbf{Models}} &
		\multirow{3}{*}{\textbf{Training Setting}} &
		\textbf{TUAB} &
		\textbf{ADNI} &
		\textbf{CamCAN} \\
		\cmidrule(lr){3-5}
		& & \textbf{EEG} & \textbf{fMRI} & \textbf{MEG} \\
		\cmidrule(lr){3-5}
		& & BAC (\%) $\uparrow$ &  BAC (\%) $\uparrow$ & MAE $\downarrow$ \\
		\midrule
		
		\multirow{3}{*}{Brain-OF Base}
		& Pretrained + Full FineTuning
		& 81.88 $\pm$ 0.17
		& 68.23 $\pm$ 2.07
		& 8.99 $\pm$ 0.14 \\
		
		& Pretrained + Linear Probing
		& 79.14 $\pm$ 0.24
		& 65.22 $\pm$ 1.47
		& 10.92 $\pm$ 0.22 \\
		
		& No Pretraining + Full FineTuning
		& 77.59 $\pm$ 0.27
		& 63.39 $\pm$ 2.36
		& 9.69 $\pm$ 1.07 \\
		\midrule
		
		\multirow{3}{*}{Brain-OF Huge}
		& Pretrained + Full FineTuning
		& 82.87 $\pm$ 0.18
		& 71.79 $\pm$ 2.19
		& 7.87 $\pm$ 0.22 \\
		
		& Pretrained + Linear Probing
		& 81.92 $\pm$ 0.21
		& 70.76 $\pm$ 1.22
		& 9.02 $\pm$ 0.11 \\
		
		& No Pretraining + Full FineTuning
		& 80.82 $\pm$ 0.18
		& 64.85 $\pm$ 1.03
		& 8.59 $\pm$ 0.94 \\
		
		\bottomrule
	\end{tabular}
	\end{small}
	\vskip -0.2in
\end{table}

\section{Ablation Study on Key Components}
\vskip -0.1in
To verify the contribution of key architectural components, we conduct ablation studies on the Sparse MoE and DINT attention mechanism. Specifically, we pretrain two variants: (i) replacing Sparse MoE with a dense Feed-Forward Network (FFN), and (ii) substituting DINT attention with vanilla self-attention.
As shown in Table~\ref{ablation_study_dense}, replacing Sparse MoE with a dense FFN results in the largest performance degradation across all tasks, with particularly severe drops on fMRI tasks. This suggests that a shared dense parameter space struggles to disentangle heterogeneous modality semantics, as the larger EEG/MEG corpora may dominate gradient updates in the dense model, hindering learning of subtle features from the smaller fMRI corpus. In contrast, Sparse MoE enables expert specialization, mitigating negative transfer and preserving performance even on under-represented modalities.
Replacing DINT attention with vanilla self-attention also leads to consistent but smaller performance declines (e.g., SEED-V Cohen’s $\kappa$ drops from 23.98\% to 22.82\%). This indicates that standard attention is susceptible to the low signal-to-noise ratio of neural recordings, attending to irrelevant fluctuations rather than meaningful patterns. DINT alleviates this issue by suppressing irrelevant context and emphasizing salient signals.
Overall, these ablations demonstrate that Sparse MoE and DINT attention are crucial for handling semantic heterogeneity and noise in brain signals.

\begin{table*}[h]
	\vskip -0.15in
	\caption{Ablation analysis of key components.}
	\label{ablation_study_dense}
	\centering
		\setlength{\tabcolsep}{10pt}
		\begin{small}
			\begin{tabular}{lccccc}
				\toprule
				\multirow{5}{*}{\textbf{Methods}} & \textbf{SEED-V} & \textbf{TUAB} & \textbf{ADNI} & \textbf{ADHD-200} &  \textbf{CamCAN}  \\
				\cmidrule(r{0.02in}){2-6}
				& \textbf{EEG} & \textbf{EEG} & \textbf{fMRI} & \textbf{fMRI} &  \textbf{MEG} \\
				\cmidrule(r{0.02in}){2-6}
				
				& Kappa(\%) $\uparrow$ & BAC(\%) $\uparrow$ & BAC(\%) $\uparrow$ & BAC(\%) $\uparrow$ &  MAE $\downarrow$ \\
				
				\midrule
				\textbf{Brain-OF (Base)} &
				\textbf{23.98$\pm$0.29} & 
				\textbf{81.88 $\pm$0.17} & \textbf{68.23$\pm$2.07} & 62.57$\pm$1.19 &  \textbf{8.99$\pm$0.14} \\
				\midrule
				w/ Dense FFN & 20.86$\pm$0.28 & 81.86$\pm$0.08 & 64.60$\pm$3.32& 59.10$\pm$1.80 &9.78$\pm$0.27 \\
				w/ Vanilla Attn. & 22.82$\pm$0.38 & 81.85$\pm$0.09 & 64.32$\pm$2.82 & \textbf{62.82$\pm$1.76} & 9.50$\pm$0.53
				\\
				
				\bottomrule
			\end{tabular}
		\end{small}
	\vskip -0.2in
\end{table*}

\section{Robustness of DINT Attention under Low SNR}
\vskip -0.1in
To explicitly validate the theoretical capability of DINT attention to handle low signal-to-noise ratio (SNR) conditions, we conducted a controlled SNR stress test on the TUAB dataset using Brain-OF (Base). We systematically injected varying levels of noise (from a high SNR of 20\,dB down to a severe -5\,dB) into the evaluation signals. As shown in Table~\ref{tab:snr_stress}, under high SNR (20\,dB), the performance difference between DINT and vanilla attention is modest (+1.25\%). However, as the noise level becomes extreme (-5\,dB), vanilla attention suffers a catastrophic representation collapse (dropping to 63.96\%). In contrast, DINT attention demonstrates strong robustness, maintaining 69.29\% (+5.33\%). These results provide empirical evidence that the differential/integral mechanism improves robustness under low-SNR conditions by stabilizing attention against noise.

\begin{table*}[h]
	\vskip -0.1in
	\caption{SNR stress test on the TUAB dataset comparing DINT and vanilla attention.}
	\label{tab:snr_stress}
	\centering
		\setlength{\tabcolsep}{10pt}
		\begin{small}
			\begin{tabular}{lccccc}
				\toprule
				\textbf{Methods} & \textbf{20\,dB} & \textbf{10\,dB} & \textbf{5\,dB} & \textbf{0\,dB} &  \textbf{-5\,dB} \\
				
				\midrule
				DINT Attention &	\textbf{80.18$\pm$0.33} &	\textbf{75.27$\pm$0.88} &	\textbf{75.02$\pm$0.28} &	\textbf{73.47$\pm$1.11} &	\textbf{69.29$\pm$2.34} \\
				Vanilla Attention &	78.93$\pm$0.40 & 74.20$\pm$1.06 & 72.71$\pm$1.21 & 70.53$\pm$1.86 &	63.96$\pm$3.47\\
				\bottomrule
			\end{tabular}
		\end{small}
	\vskip -0.2in
\end{table*}

\section{Parameter-Efficient Fine-Tuning}
\vskip -0.1in
While our primary downstream evaluations adopt full fine-tuning to fully exploit model capacity, this approach incurs substantial memory costs. To enable more efficient adaptation of Brain-OF, we explore Low-Rank Adaptation (LoRA) \cite{hu2022lora}, a widely used parameter-efficient fine-tuning (PEFT) method.
Specifically, we freeze the pretrained backbone and introduce trainable low-rank decomposition matrices (rank $r=32$, scaling factor $\alpha=64$), updating only $\sim$2\% of the total parameters. As shown in Table~\ref{lora}, LoRA achieves competitive performance compared to full fine-tuning. Notably, on small-scale fMRI datasets, LoRA slightly outperforms full fine-tuning (e.g., ADNI: 68.23\% $\rightarrow$ 68.44\%, LEMON-fMRI: 58.95\% $\rightarrow$ 59.36\%). This suggests that the constrained parameter space of LoRA acts as an implicit regularizer, mitigating overfitting in low-data regimes. Although a modest performance drop is observed on more complex tasks (e.g., SEED-V), these results demonstrate that PEFT provides a practical and efficient alternative for adapting Brain-OF, especially in low-resource settings.

\begin{table*}[h]
	\vskip -0.13in
	\caption{Comparison between full fine-tuning and LoRA-based parameter-efficient fine-tuning on Brain-OF (Base).}
	\label{lora}
	\centering
	\setlength{\tabcolsep}{4.3pt}
	\begin{small}
		\begin{tabular}{lcccccc}
			\toprule
			\multirow{5}{*}{\textbf{Methods}} & \textbf{SEED-V} & \textbf{TUAB} & \textbf{ADNI} & \textbf{ADHD-200} &  \textbf{CamCAN} & \textbf{LEMON} \\
			\cmidrule(r{0.02in}){2-7}
			& \textbf{EEG} & \textbf{EEG} & \textbf{fMRI} & \textbf{fMRI} &  \textbf{MEG} & \textbf{fMRI} \\
			\cmidrule(r{0.02in}){2-7}
			
			& Kappa(\%) $\uparrow$ & BAC(\%) $\uparrow$ & BAC(\%) $\uparrow$ & BAC(\%) $\uparrow$ &  MAE $\downarrow$ & BAC(\%) $\uparrow$ \\
			
			\midrule
			Full Finetuning &
			\textbf{23.98$\pm$0.29} & 
			\textbf{81.88 $\pm$0.17} & 68.23$\pm$2.07 & \textbf{62.57$\pm$1.19} &  \textbf{8.99$\pm$0.14} & 58.95±2.13 \\
			LoRA ($\sim$2\% Params) & 21.81$\pm$0.13	& 81.30$\pm$0.04 & \textbf{68.44$\pm$2.13} & 62.10$\pm$1.04 & 9.17$\pm$0.44 & \textbf{59.36$\pm$3.02}\\
			\bottomrule
		\end{tabular}
		\begin{minipage}{\linewidth}
			\footnotesize
			LoRA updates only $\sim$2\% of parameters while maintaining competitive performance, and even improves results on small-scale fMRI datasets, indicating enhanced robustness in low-data regimes.
		\end{minipage}
	\end{small}
	\vskip -0.2in
\end{table*}

\section{Label Efficiency and Low-Resource Analysis}
\begin{figure}[h]
	\vskip -0.05in
	\centering
		\includegraphics[width=\columnwidth]{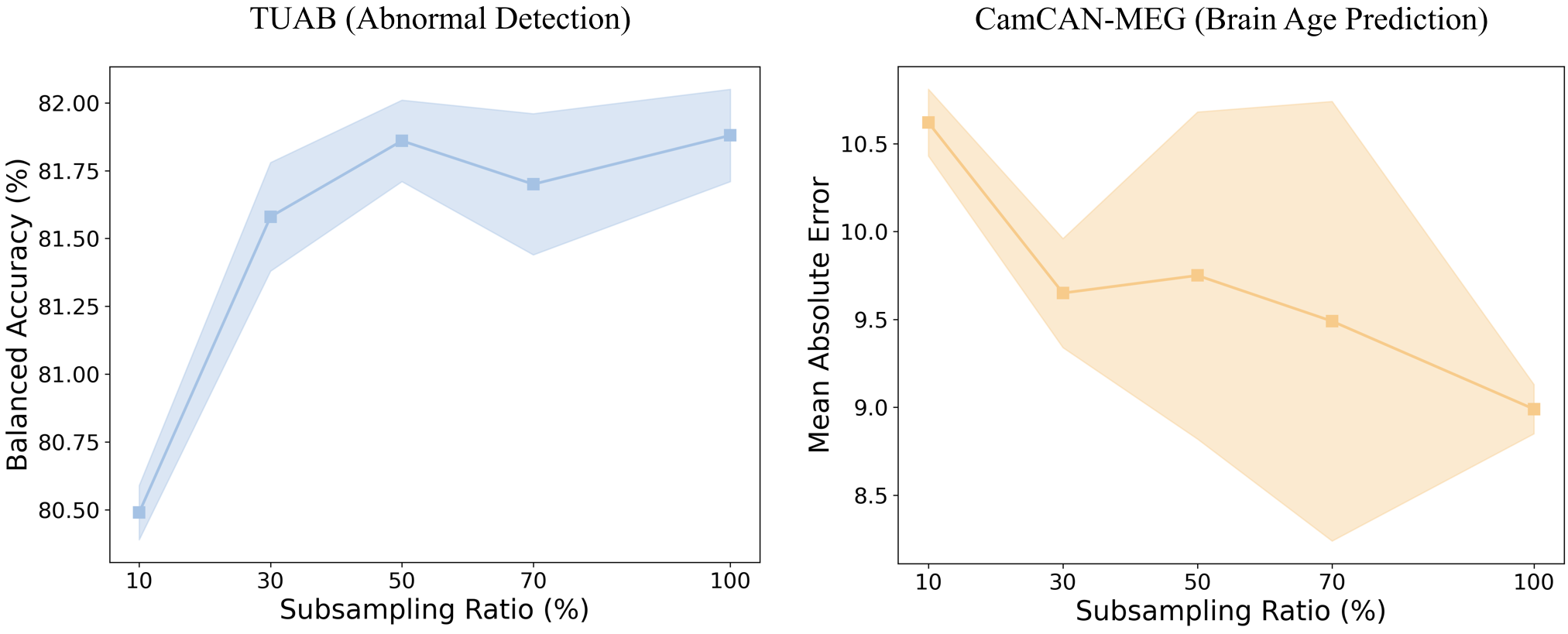}
		\vskip -0.05in
		\caption{
			\textbf{Label efficiency of Brain-OF under limited supervision.} Brain-OF Base is evaluated across varying training data fractions (10\%, 30\%, 50\%, 70\%, 100\%).
			Shaded regions denote standard deviation across five random seeds.
			Brain-OF achieves strong performance even with limited labeled data, highlighting the effectiveness of heterogeneous multimodal pretraining for low-resource settings.
		}
		\label{label_efficiency}
\end{figure}

\vskip -0.05in
To evaluate Brain-OF’s robustness under data-scarce conditions, we assess the Brain-OF Base using progressively smaller fractions (10\%–100\%) of the downstream training data. We consider two representative tasks: abnormality detection on TUAB (EEG) and brain-age prediction on CamCAN (MEG).
As illustrated in Figure~\ref{label_efficiency}, Brain-OF exhibits strong label efficiency across both tasks. For TUAB (left), the model achieves over 81.5\% balanced accuracy using only 30\% of the labeled data, nearly approaching the full-data performance of 81.88\%. For CamCAN (right), the mean absolute error consistently decreases as more data becomes available, yet the model maintains competitive accuracy even in low-resource settings (e.g., 30\% training data).
These results indicate that the rich, generalizable representations learned during large-scale multimodal pretraining reduce the dependency on large labeled downstream datasets. Brain-OF thus offers promising potential for low-resource neuroimaging scenarios, where annotated data is often limited or expensive to obtain.

\section{Limitations and Future Work}
\label{limitations}
Although we curate one of the largest neuroimaging corpora to date, which spans over 40 datasets across fMRI, EEG and MEG, the overall data scale remains orders of magnitude smaller than that of foundation models in natural language processing or computer vision. Moreover, the pretraining corpus exhibits notable modality imbalance, with fMRI samples being considerably scarcer than EEG and MEG recordings. Future iterations of Brain-OF may address these challenges by incorporating additional modalities such as fNIRS and PET, and by extending the framework to support more general multivariate time series.

Currently, Brain-OF employs a relatively straightforward multimodal fusion strategy. While effective, this approach may not fully capture complex and nonlinear cross-modal interactions. Future work could explore more expressive fusion mechanisms, such as adaptive gating strategies or dedicated cross-modal attention modules, to better model fine-grained inter-modality dependencies.


\section{Broader Impacts}
\label{broader_impacts}
This work presents Brain-OF, an omnifunctional foundation model jointly pretrained on fMRI, EEG and MEG, representing a significant advance toward unifying heterogeneous neuroimaging modalities within a single framework. Compared to single-modality foundation models, Brain-OF exhibits potentially stronger generalization capabilities, even to unseen or specialized functional modalities.
Consequently, Brain-OF has a broader impact on multimodal fusion in neuroscience research and clinical applications, empowering researchers to integrate the high spatial resolution of fMRI with the temporal precision of EEG/MEG for more accurate diagnosis in epilepsy, neurodegenerative diseases and cognitive disorders.
Moreover, by transferring knowledge from capital-intensive modalities (fMRI/MEG) to accessible, wearable ones (EEG), this work accelerates the deployment of high-performance neurotechnologies in real-world and portable settings.
We release Brain-OF Huge, one of the largest open-source brain foundation models to date. This provides a reliable, pre-validated backbone for the neuroscience and brain-computer interface community, substantially lowering the barrier to entry for researchers who lack resources to pretrain large-scale models from scratch. By enabling the reuse of rich, cross-modal representations, we aim to broaden access to powerful neural representations and foster a collaborative, data-centric ecosystem for future neuroimaging research.


\end{document}